\documentclass[review]{elsarticle}

\usepackage{microtype}
\usepackage{amsmath}
\usepackage{amssymb}
\usepackage{graphicx}
\usepackage{setspace}
\usepackage{subfigure}
\usepackage{booktabs} 
\usepackage{amssymb}

\usepackage{algorithm}
\usepackage{algorithmic}
\usepackage{lineno,hyperref}

\modulolinenumbers[5]

\journal{Journal of arXiv}

\begin{document}
	
	\begin{frontmatter}
		
		\title{Regularly Updated Deterministic Policy Gradient Algorithm}
		\author[mymainaddress,mysecondaryaddress]{Shuai Han}
		
		\author[mymainaddress,mysecondaryaddress]{Wenbo Zhou}
		
		\author[mymainaddress,mysecondaryaddress]{Shuai L\"u\corref{mycorrespondingauthor}}
		
		\ead{lus@jlu.edu.cn}
		
		\author[mymainaddress,address3]{Jiayu Yu}

		\cortext[mycorrespondingauthor]{Corresponding author}
		
		\address[mymainaddress]{Key Laboratory of Symbolic Computation and Knowledge Engineering (Jilin University), Ministry of Education, Changchun 130012, China}
		\address[mysecondaryaddress]{College of Computer Science and Technology, Jilin University, Changchun 130012, China}
		\address[address3]{College of Software, Jilin University, Changchun 130012, China}

		\begin{abstract}
			Deep Deterministic Policy Gradient (DDPG) algorithm is one of the most well-known reinforcement learning methods. However, this method is inefficient and unstable in practical applications. On the other hand, the bias and variance of the Q estimation in the target function are sometimes difficult to control. This paper proposes a Regularly Updated Deterministic (RUD) policy gradient algorithm for these problems. This paper theoretically proves that the learning procedure with RUD can make better use of new data in replay buffer than the traditional procedure. In addition, the low variance of the Q value in RUD is more suitable for the current Clipped Double Q-learning strategy. This paper has designed a comparison experiment against previous methods, an ablation experiment with the original DDPG, and other analytical experiments in Mujoco environments. The experimental results demonstrate the effectiveness and superiority of RUD.
		\end{abstract}
		
		
	\end{frontmatter}

	\section{Introduction}
	
	\label{Introduction}
	
	Reinforcement learning algorithms based on deterministic policy gradient are widely used in various complex tasks \cite{qiu2019deep}\cite{kim2019optimizing}\cite{yang2019application}\cite{le2018learning} because their policy gradients are easy to estimate \cite{silver2014deterministic} and they can effectively solve problems with high-dimensional action spaces \cite{lillicrap2015continuous}. However, although such methods have been successfully applied in applications, algorithms based on deterministic policy gradient are still notoriously inefficient \cite{islam2017reproducibility}, unstable \cite{khadka2018evolution}\cite{pourchot2018cem}, and the convergence is sensitive to parameters. Therefore, how to improve deterministic policy gradient methods remains a hotspot in this field.
	
	The perspectives of improvement for deterministic policy gradient methods includes: 
	
	\begin{itemize}
		\item Exploration, such as variational information
		maximization\cite{houthooft2016vime}, count-based exploration\cite{ostrovski2017count} \cite{tang2017exploration}, intrinsic motivation \cite{bhatnagar2009convergent} and curiosity \cite{pathak2017curiosity};
		\item Data utilization, such as DDPG based on prioritized experience replay \cite{schaul2015prioritized} \cite{horgan2018distributed}. 
	\end{itemize}

	Although the above two methods have achieved excellent results in specific fields \cite{fortunato2017noisy} \cite{khadka2018evolution}, the former usually involves a complex structure, while the latter needs to regulate the parameters of the importance sampling part. Improvement methods on the target value calculation and network update of deterministic policy gradient methods, such as Twin Delayed Deep Deterministic (TD3) policy gradient algorithm\cite{fujimoto2018addressing} and smoothie\cite{nachum2018smoothed}, fundamentally overcome the inherent deficiencies of DDPG when calculating the target, but still do not solve the low data utilization problem brought by the procedure of classical reinforcement learning methods.
	
	This paper proposes Regularly Updated Deterministic (RUD) policy gradient algorithm. RUD mitigates the inherent shortcomings of data utilization in the classical off-policy reinforcement learning methods by allowing the exploration and the learning process to be performed alternately and concentratedly. First, we theoretically analyze characteristics of the data utilization under the classical off-policy reinforcement learning paradigm, and pointed out the difficulties for the agent to sample new experience. Then, we present the RUD algorithm and prove that RUD can better use new experiences than the classical algorithm. In addition, for the Clipped Double Q-learning method used in TD3, we theoretically prove the existence of downward deviation as well as its magnitude, and empirically explain why  the procedure of RUD can mitigate this deviation. Finally, we confirm the superiority of RUD through a series of experiments by comparing with previous methods, exploring parameter settings and analyzing efficiency of data utilization in RUD.
	
	The rest of this paper is organized as follow. Section 2 introduces some related work. Section 3 presents the background; Section 4 analyzes the insufficient use of new experiences in the procedure of classical algorithm, and proposes RUD, and theoretically proves that RUD can better use of the new experiences. Section 5 analyzes deviation of Clipped Double Q-learning, and explains that RUD's procedure can better alleviate this deviation; Section 6 conducts a series of experiments to prove the superiority of RUD. Section 7 summarizes the work of this paper and discusses the future work.
	
	\section{Related Works}
	\label{Related Work}
	
	The main contribution of this paper is the design and development of RUD which allocates the exploration and exploitation process alternately and concentratedly. In RUD, the replay buffer introduces new experiences in blocks, thereby improving the utilization of new experiences. Therefore, the work of this paper can be regarded as an improvement in the data utilization ability of deterministic policy gradient algorithms. There are many previous works with regard to improving the data utilization of reinforcement learning. Experience replay breaks the temporal correlations by uniformly sampling more and less recent experience for the updates \cite{lin1992self}. Subsequent researches also show that off-policy reinforcement learning has higher sample efficiency \cite{mnih2015human} \cite{lillicrap2015continuous} \cite{haarnoja2018soft}. Prioritizing the data in the replay buffer can further improve the sample efficiency of off-policy reinforcement learning, such as the priority scanning method according to the next state to be updated \cite{moore1993prioritized} \cite{andre1998generalized}, using TD error to prioritize experience in model-based methods \cite{van2013planning} or model-free methods \cite{schaul2015prioritized}.
	
	In the RUD algorithm paradigm, the policy is sufficiently perturbed because the exploration is carried out concentratedly. Therefore, RUD can also be seen as an exploration improvement for classical reinforcement learning methods. Previous methods for exploration in reinforcement learning include: variational information
	maximization\cite{houthooft2016vime}, count-based exploration\cite{ostrovski2017count}\cite{tang2017exploration}, intrinsic motivation \cite{bhatnagar2009convergent} and curiosity \cite{pathak2017curiosity}. In addition, noisy methods have been used for the exploration of parameter space \cite{fortunato2017noisy} \cite{plappert2017parameter}. Evolutionary reinforcement methods also use mutation strategies in populations to explore parameter space \cite{khadka2018evolution} \cite{pourchot2018cem} \cite{lu2019recruitment}.
	
	Our method is based on the Deterministic Policy Gradient (DPG) algorithm \cite{silver2014deterministic} and combines advanced optimization techniques used by DDPG \cite{lillicrap2015continuous} and TD3 \cite{fujimoto2018addressing}. In TD3, the Clipped Double Q-learning technology is used to alleviate the overestimation problem, and we also analyzed the problem of underestimation introduced in TD3. Other work related to DPG includes model-free algorithms using model-based methods \cite{popov2017data},  multi-step returns and prioritized experience replay \cite{horgan2018distributed}, and distribution-based methods \cite{barth2018distributed}.
	
	\section{Background}
	\label{Background}
	
	This section introduces the deterministic policy gradient algorithm and the optimization of it.
	
	\subsection{Deterministic Policy Gradient Algorithm}
	\label{Deterministic Policy Gradient}
	
	The interaction between an agent and a continuous environment can be modeled as a Markov Decision Process (MDP). This process includes a continuous state space $S$, a continuous action space $A$, an initial state probability density $p(s_1)$, an unknown transition probability density $p(s_{t+1}|s_{t},a_{t})$ and a reward function $r$: $S \times A \rightarrow \mathbb{R}$. The transition probability density satisfies the Markov property $p(s_{t+1}|s_{1}, a_{1}, ..., s_{t}, a_{t}) = p(s_{t+1}|s_{t},a_{t})$ for any trajectory $s_{1}, a_{1}, ..., s_{t}, a_{t}$. A policy is used to select actions in MDP. A deterministic policy can be denoted as: $\mu_{\theta}: S \rightarrow A$, where $\theta \in \mathbb{R}^{n}$ is a vector of $n$ parameters. $R_t = \sum_{k=1}^\infty \gamma^kr_{t+k}$ denotes the total discounted reward from the time-step $t$, where $\gamma \in (0, 1]$ is a discount factor.
	
	The goal of an agent is to adjust its policy to maximize the expected total discounted return $\mathbb{E}[R_1|\mu_{\theta}]$. The probability density of reaching $s'$ from $s$ conducting policy $\mu_{\theta}$ after $t$ time steps can be denoted as $p(s\rightarrow s', t, \mu_{\theta})$. The discounted state distribution can be denoted as: $\rho^{\mu_{\theta}}(s') = \int_{S}\sum_{t=1}^\infty\gamma^{t-1}p_1(s)p(s\rightarrow s', t, \mu_{\theta})ds$. Then the performance objective of an agent can be denoted as:
	$$
	\label{object}
	J(\mu_{\theta}) = \int_{S}\rho^{\mu_{\theta}}(s)r(s, \mu_{\theta}(s))ds = \mathbb{E}_{s\sim\rho^{\mu_{\theta}}}[r(s, \mu_{\theta}(s))]  \eqno(1)
	$$
	Silver et al. proved that the deterministic policy gradient in Eq. (1) exists under certain conditions \cite{silver2014deterministic} \cite{lillicrap2015continuous}. The deterministic policy gradient can be denoted as:
	$$
	\label{object gradient}
	\nabla_{\theta} J(\mu_{\theta})=\mathbb{E}_{s\sim\rho^{\mu_{\theta}}}[\nabla_{\theta}\mu_{\theta}(s)\nabla_{a}Q^{\mu_{\theta}}(s, a)|_{a=\mu_{\theta}(s)}]  \eqno(2)
	$$
	where $Q^{\mu_{\theta}}(s, a) = \mathbb{E}_{s\sim\rho^{\mu_{\theta}},a\sim \mu_{\theta}}[R_t|s,a]$. The $Q$ function is learned by minimizing the critic loss: $L = (y - Q(s, a))^2$, where the target value $y = r + \gamma Q(s', a')$.

	\subsection{Advanced Optimization for Deterministic Policy Gradient}
	\label{Advanced Optimization for Deterministic Policy Gradient}
	
	Deep deterministic policy gradient (DDPG) algorithm is a widely used deterministic policy gradient algorithm. DDPG creates target actor and critic networks to calculate the target value: $y = r+\gamma Q'(s',\mu'(s'|\theta^{\mu'})|\theta^{Q'})$. $\theta^{Q'}$ and $\theta^{\mu'}$ is updated by soft updating: $\theta^{Q'} = (1 - \tau)\theta^{Q'} + \tau \theta^{Q}$, $\theta^{\mu'} = (1 - \tau)\theta^{\mu'} + \tau \theta^{\mu}$, where $\tau << 1$. DDPG also applies the batch normalization technique \cite{ioffe2015batch} to calculate gradients and an Ornstein-Uhlenbeck process \cite{uhlenbeck1930theory} to execute exploration \cite{lillicrap2015continuous}.
	
	Twin Delayed Deep Deterministic (TD3) policy gradient algorithm is the state-of-art deep deterministic policy gradient method. TD3 has made several optimizations for the original deep deterministic policy gradient algorithm. In TD3, the target value is calculated by: $y=r+\gamma min_{i=1,2}Q'_{i}(s',\pi(s'|\theta^{\mu'})+\epsilon|\theta^{Q'})$, where $\epsilon\sim clip(N(0,\sigma),-c,c)$. In order to solve the overestimation problem in DDPG, TD3 updates the target value by taking the minimum value of two independent $Q$ networks. Although this way introduces the underestimation problem, it alleviates the overestimation of the target value. TD3 introduces a regularization method by adding a small amount of noise to the target policy to smooth the value estimation, which can optimize the learning of the value function. In addition, TD3 uses Delayed Policy Updates to optimize the update of policy networks and value networks. It is considered that this approach can reduce the variance of the value function estimation \cite{fujimoto2018addressing}. Both DDPG and TD3 apply the process shown in Algorithm \ref{alg:DPG}. \begin{algorithm}[h]
		\caption{Classical DPG}
		\label{alg:DPG}
		\begin{algorithmic}
			\STATE {\bfseries Input:} Batch size $N$, Learning parameters $params$
			\STATE Initialize actor and critic networks
			\STATE Initialize replay buffer $R$
			\FOR{$t=1$ {\bfseries to} $T$}
			\STATE Select action according to exploration strategy
			\STATE Store the transition tuple $(s, a, r, s')$ into $R$
			\IF{$t \geq N$}
			\STATE Sample mini-batch of $N$ transitions $(s, a, r, s')$ from $R$
			\STATE Update critics according to the critic loss and $params$
			\STATE Update actors according to the policy gradient and $params$
			\ENDIF
			
			\ENDFOR
		\end{algorithmic}
	\end{algorithm}

	\section{Regularly Updated Deterministic Policy Gradient Algorithm}
	
	This section theoretically points out the problem that it is difficult for the agent to sample new experience under the classical process as Algorithm \ref{alg:DPG}. Then we give the RUD algorithm, and prove that RUD can use the new experience more frequently than the classical algorithm.
	
	\subsection{Data Exploitation of Classical DPG}
	
	Although DDPG, TD3 and other deterministic policy gradient algorithms are different in learning details, they are all designed according to the ``learning while exploring" paradigm, as shown in Algorithm \ref{alg:DPG}. In this paradigm,  the experiences are sampled uniformly before calculating the gradient and updating the networks at each time step, and the size of the replay buffer increases at every time step. This results in that the earlier the experience is stored, the more times it is sampled, and the later the experience is stored, the fewer times it is sampled. In this subsection, we will theoretically discuss the frequency difference of being sampled among the experiences stored at different time steps in Algorithm \ref{alg:DPG}, and empirically point out the impact of this difference on the data utilization of deterministic policy gradient algorithms.

	For convenience, we assume that the size of the replay buffer is equal to the total time budget $T$ of the learning process. This assumption is only to ensure that the size of the replay buffer is increasing during the entire learning process. In fact, in experiments settings of many deterministic policy gradient algorithms, the size of the replay buffer is indeed equal to the total time step \cite{fujimoto2018addressing} \cite{lu2019recruitment}. Even if the total time budget $T$ is larger than the size of the replay buffer, the stage that the size of the replay buffer is increasing will always exist in the early learning process.
	
	\newtheorem{thm}{\bf Theorem}
	\begin{thm}\label{thm1}
		Suppose that $M_t$ denotes the number of times that experience stored at the $t$ time step is played back in the subsequent learning process. Then $E[M_{t_1}]> E[M_{t_2}]$ if $N \leq t_1 < t_2 \leq T$, where $N$ is the batch-size of samples. The maximum value of $E[M_t]$ is $ Nln(T+1/N)$, the minimum value is $N/T$.
	\end{thm} 
	
	{{\noindent\it Proof:}\quad}
	{\it There exists a sequence that $\{a_1, a_2,..., a_T\}$, where $a_t = E[M_{t}], t=1,2,...,T$. Suppose that $M_{t}^{t'}$ denotes the number of times that the experiences stored at the $t$ time step are played back at the $t'$ time step. Because of that $M_{t} = M_{t}^{t} + M_{t}^{t+1} + M_{t}^{t+2} + ... +M_{t}^{T}$,  when $N \leq t < T$, $a_t = E[M_{t}] = E[M_{t}^{t}] +E[M_{t}^{t+1}] + E[M_{t}^{t+2}] + ... + E[M_{t}^{T}]$. And $a_{t+1} = E[M_{t+1}]=E[M_{t+1}^{t+1}] + E[M_{t+1}^{t+2}] + ... + E[M_{t+1}^{T}] =E[M_{t}^{t+1}] + E[M_{t}^{t+2}] + ... + E[M_{t}^{T}]$. Therefore $ a_t - a_{t+1} = E[M_{t}^{t}] > 0$. Then $\{a_{N}, a_{N+1},..., a_T\}$ is a decreasing sequence. Then $E[M_{t_1}] > E[M_{t_2}]$ if $N \leq t_1 < t_2 \leq T$.
		
		Due to uniform sampling and sampling once at each time step for Algorithm \ref{alg:DPG}, $E[M_{t}^{t'}] = C_1^1C_{t'-1}^{N-1}/C_{t'}^{N}$, where $1 \leq t \leq t' \leq T$. Then $E[M_{t}] = E[M_{t}^{t}] +E[M_{t}^{t+1}] + E[M_{t}^{t+2}] + ... + E[M_{t}^{T}] = C_1^1C_{t-1}^{N-1}/C_{t}^{N} + C_1^1C_{t}^{N-1}/C_{t+1}^{N} + C_1^1C_{t+1}^{N-1}/C_{t+2}^{N} + ... + C_1^1C_{T-1}^{N-1}/C_{T}^{N} = N/t + N/(t+1) + N/(t+2) + ... + N/T$. In Algorithm \ref{alg:DPG}, the agent does not perform the sampling process when $t<N$. Therefore, the experiences stored in the first $N$ time steps will be played back with the same number of times in the subsequent learning process: $E[M_{1}] = E[M_{2}] = ... = E[M_{N}]$. And $\{E[M_{N}], E[M_{N+1}],..., E[M_{T}]\}$ is a decreasing sequence. So sequence  $\{E[M_{i}]\}, i = 1, 2, ..., T$, $E[M_{i}]$ takes the maximum value when $i = 1, 2 ,..., N$. The maximum value is $E[M_{i}] = E[M_{N}] = N/N + N/(N+1) + N/(N+2) + ... + N/T$. Let $S(n) = 1/1 + 1/2 + ... + 1/n$, $S(n) = ln(n+1) + C$, then $E[M_{i}] = N(1/N + 1/(N+1) + 1/(N+2) + ... + 1/T) = N(S(T) - S(N-1)) = Nln((T+1)/N)$. $E[M_{i}]$ takes the minimum value when $i = T$. The minimum value is $N/T$.
		$\hfill\blacksquare$ 
		
	}
	
	Theorem \ref{thm1} shows that in the classical DPG algorithm paradigm, the expectation of the number of times, that experiences entering the replay buffer at different time steps is played back, is different, This difference of expectation may be very large. Assuming that in a training process, the total time step $T$ is $1,000,000$ and the batch-size $N$ is $128$, then according to Theorem \ref{thm1}, the expected number of times of replaying the first experience batch stored in the buffer is $E[M_{128}] = 1,147.33$,  and the expected number of times of replaying the last experience stored in the replay buffer is $E[M_{1,000,000}] = 0.000128$. This difference may lead to the excessive use of the first batch of experience in the replay buffer. At the same time, the last batch of experience stored in the replay buffer may not be learned because it is rarely or even not played back in Algorithm \ref{alg:DPG}.
	
	Theorem \ref{thm1} also can be elaborated as that the agent is more inclined to learn old experiences rather than new experiences in the classical DPG procedure. However, on the one hand, old experiences are usually already learned. On the other hand, new experiences are more likely to include unexperienced states and unlearned knowledge about the environment or reward function. In the next subsection, we will introduce a new DPG algorithm paradigm, which allows agents to learn more about new experiences while retaining uniform sampling.

	\subsection{Better Procedure for DPG}
	
	The classical DPG algorithm learns old experiences more, but always replays new experiences less. This is caused by its ``exploring while learning" mode. In this subsection, we will propose Regularly Updated Deterministic policy gradient algorithm. The procedure of this algorithm is shown in Algorithm \ref{alg:our DPG}.\begin{algorithm}[h]
		\caption{Regularly Updated DPG}
		\label{alg:our DPG}
		\begin{algorithmic}
			\STATE {\bfseries Input:} Batch size $N$, learning parameters $params$, fixed size $F$
			\STATE Initialize actor and critic networks
			\STATE Initialize replay buffer $R$
			\STATE t = 0
			\WHILE {$t \leq T$}
			\FOR{$f=1$ {\bfseries to} $F$}
			\STATE Select action according to exploration strategy
			\STATE Store the transition tuple $(s, a, r, s')$ into $R$
			\ENDFOR
			\FOR{$f=F$ {\bfseries to} $1$}
			\IF{$ N \leq t \leq ？T$}
			\STATE Sample mini-batch of $N$ transitions $(s, a, r, s')$ from $R$
			\STATE Update critics according to the critic loss and $params$
			\STATE Update actors according to the policy gradient and $params$
			\ENDIF
			\ENDFOR
			\STATE $t = t + F$
			\ENDWHILE 
		\end{algorithmic}
	\end{algorithm} Regularly Updated Deterministic policy gradient algorithm uses the separated exploration and learning process.
	The agent first conducts $F$ time steps to interact with the environment and generate experiences. Then the agent performs $F$ time steps to sample experiences and update parameters. Such a process can make exploration and learning alternately concentrated. Intuitively speaking, RUD introduces new experiences in batches instead of individually before sampling, which makes new experiences account for a larger proportion in the replay buffer. Thereby it can increase the probability of replaying new experiences for the agent. In the rest part of this subsection, we will theoretically prove that RUD can make more use of new experiences than the classical DPG algorithm.

	\begin{thm}\label{thm2}
		Suppose that $M^{(1)}(t:F)$ and $M^{(2)}(t:F)$ represent the number of times that the latest stored $F$ experiences are sampled by Algorithm \ref{alg:DPG} and Algorithm \ref{alg:our DPG} respectively in the $F$ times sampling before $t$. Then,  $E[M^{(1)}(t:F)] < E[M^{(2)}(t:F)]$.
	\end{thm} 
	
	{{\noindent\it Proof:}\quad}
	{\it 	
		According to proof process of Theorem \ref{thm1}, $E[M^{(2)}(t:F)] = (1-(C_{t-F}^{N}/C_{t}^{N}))*F = 1-(C_{t-F}^{N}/C_{t}^{N}) + 1-(C_{t-F}^{N}/C_{t}^{N}) + ... + 1-(C_{t-F}^{N}/C_{t}^{N}) $, 
		and $E[M^{(1)}(t:F)] = 1-(C_{t-F+1}^{N}/C_{t}^{N}) + 1-(C_{t-F+2}^{N}/C_{t}^{N}) + ... + 1-(C_{t-F+F}^{N}/C_{t’}^{N}) $.
		
		Since for any $1\leq i<F$, there is $1-(C_{t-F+i}^{N}/C_{t}^{N}) < (1-(C_{t-F}^{N}/C_{t}^{N}))$, therefore $E[M^{(1)}(t:F)] < E[M^{(2)}(t:F)]$.
		$\hfill\blacksquare$ 
		
	}
	
	Theorem \ref{thm2} shows that the algorithm \ref{alg:our DPG} introduces new experiences in batches to increase the sampling proportion of recent $F$ experiences in the replay buffer, thereby increasing the number of sampling times for new experiences. 
	
	The setting of $F$ is worth considering carefully. Suppose that during a training process, $T=1,000,000$, and the size of the replay buffer is equal to $T$. If the parameter $F$ in RUD is equal to $T$,  the agent will first generate $1,000,000$ experiences during the interaction process with the environment and then perform $1,000,000$ times of sampling and updating. In this case, the expectation of the sampling number of each experience in the replay buffer is equal. So the utilization rate of new experience is higher. However, this kind of setting will also lead to lack of exploration. Because in the 1,000,000 time steps when the agent interacts with the environment, the policy network has not changed, so its exploration policy is only based on a random disturbance of the initial policy. So larger $F$ leads to lower exploration efficiency and higher data utilization. If $F$ is set to $1$, the agent's policy for exploration will change at each time step. Therefore its exploration efficiency may be relatively high. However, according to Theorem \ref{thm1}, the agent's use of new experience is insufficient under this circumstance. So smaller $F$ leads to higher exploration efficiency and lower data utilization. Therefore, the setting of the parameter $F$ is essentially a compromise between the exploration and exploitation efficiency of the agent.

	\section{More Accurate Target of RUD}
	
	In order to solve the overestimation problem in DDPG, the state-of-art method TD3 use the Clipped Double Q-learning strategy, but this strategy also introduces underestimation problem which yields a downward bias.  In this section, we first theoretically analyze the existence and size of this bias, and then empirically show that RUD learning settings can mitigate this bias.
	
	In the classical DDPG algorithm, using a single target critic network to calculate the target value will cause the overestimation problem \cite{lillicrap2015continuous} \cite{fujimoto2018addressing}. In TD3, after sampling $(s, a, r, s')$ from the replay buffer, the calculation of the target value used to update the two critical networks $Q_{\theta_1}, Q_{\theta_2}$ is: $y = r + \gamma min_{i=1,2}Q'_{i}(s',\pi(s'|\theta^{\mu'})|\theta^{Q'})$. Although this update method can alleviate the overestimation problem, it still brings the accuracy problem of the target value. We will prove that the way TD3 calculates the target value will introduce a downward deviation, and we will give the magnitude of this deviation under certain conditions.
	
	\begin{thm}\label{thm3}
		Suppose that $Q_1$ and $Q_2$ are on the whole unbiased in the sense that $\int_A(Q_1(s, a) - V_*(s)) da = 0$ and $\int_A(Q_2(s, a) - V_*(s)) da = 0$ for some $V_*(s)$. Then, when $Q_1\not=Q_2$, $E_{a\sim\mu}[min_{i = 1,2} Q_i(s, a)] < V_*(s)$. If the values of $Q_1$ and $Q_2$ follow the Gaussian distribution with the same variance $\sigma$, then $E_{a\sim\mu}[min_{i = 1,2} Q_i(s, a)] = V_*(s) - \sigma/\sqrt{\pi}$.
	\end{thm} 
	
	{{\noindent\it Proof:}\quad}
	{\it Considering that $\int_A(Q_1(s, a)-V_*(s)) da = 0$, It can be derived that $V_*(s) = \int_A Q_1(s, a) da =E_{a\sim\mu}[ Q_1(s, a)]$; Similarly, $V_*(s) = \int_A Q_2(s, a) da =E_{a\sim\mu}[Q_2(s, a)]$. Then, $E_{a\sim\mu}[min_{i=1, 2}Q_i(s, a)] = \frac{1}{2}E_{a\sim\mu}[Q_1(s, a) + Q_2(s, a) - |Q_1(s, a) - Q_2(s, a)|] = V_*(s) - \frac{1}{2}E_{a\sim\mu}[|Q_1(s, a) - Q_2(s, a)|]$. Also considering that $Q_1\not=Q_2$, $ E_{a\sim\mu}[|Q_1(s, a)-Q_2(s, a)|]> 0$, so $E_{a\sim\mu} [min_{i = 1,2} Q_i(s, a)] <V_*(s)$.
		
		When the values of $Q_1$ and $Q_2$ follow the Gaussian distribution of the same variance $\sigma$, that is, $Q_1, Q_2 \sim N(V_*(s), \sigma)$. Let $X_1 = \frac{Q_1(s, a)-V_*(s)}{\sigma}$, $X_2 = \frac{Q_2(s, a)-V_*(s)}{\sigma}$ , Then $X_1, X_2\sim N(0, 1)$, and $Q_1(s, a) = \sigma X_1 + V_*(s)$, $Q_2(s, a) = \sigma X_2 + V_*(s)$. So there is $E_{a\sim \mu}[min_{i=1, 2}Q_i(s, a)] = E[\sigma min(X_1, X_2) + V_*(s)] = V_ *(s) + \sigma E[min(X_1, X_2)] = V_*(s) + \frac{\sigma}{2} E[X_1+X_2- |X_1-X_2|]= V_*(s) + \frac{\sigma}{2}E[|X_1-X_2|)]$. Let $Y = X_1-X_2$, then  $Y\sim N(0, 2)$, so $f_Y(y) = \frac{1}{2\sqrt{\pi}}e^{-\frac{y ^2}{4}}$.
		
		$E_{a\sim\mu}[min_{i=1, 2}Q_i(s, a)] = V_*(s) - \frac{\sigma}{2}E[|Y|] = V_*(s) - \frac{\sigma}{2}\int_{ - \infty }^{ + \infty}|y|f_Y(y)dy = V_*(s) - \sigma\int_{0}^{+\infty} y\frac{1}{2\sqrt{\pi}} e^{-\frac{y^2}{4}}dy = V_*(s) - \frac{2\sigma}{\sqrt{\pi}} \int_{0}^{+\infty} \frac{y}{2} e^{-(\frac{y}{2})^2}d(\frac{y}{2}) =  V_*(s) - \frac{\sigma}{\sqrt{\pi}}$.$\hfill\blacksquare$

	}
	
\begin{figure*}[tb]	
		\centering
		\subfigure[Hopper-v2]{
			\begin{minipage}[t]{0.5\linewidth}
				\centering
				\includegraphics[width=2.5in]{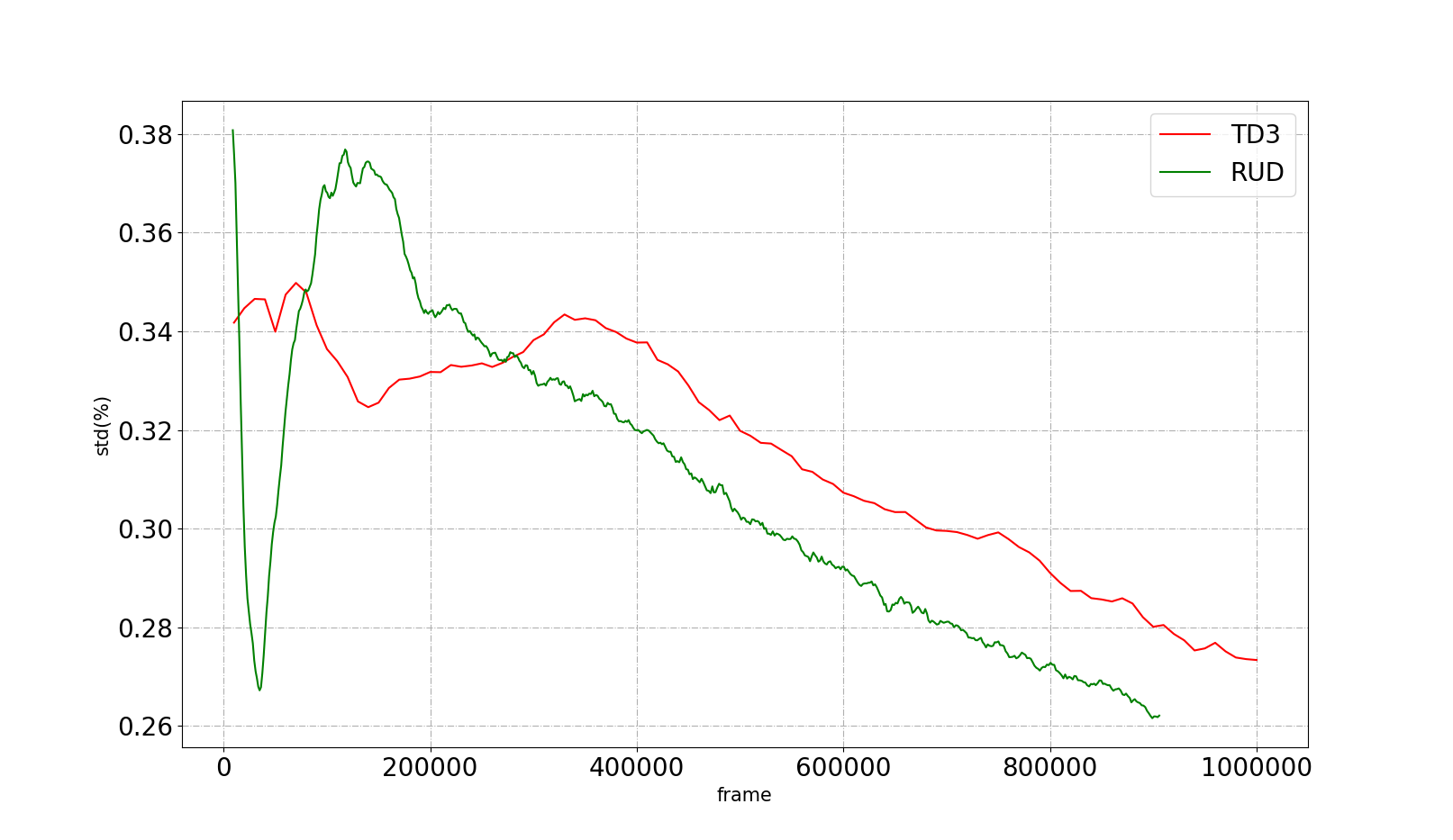}
			\end{minipage}%
		}%
		\subfigure[Walker2d-v2]{
			\begin{minipage}[t]{0.5\linewidth}
				\centering
				\includegraphics[width=2.5in]{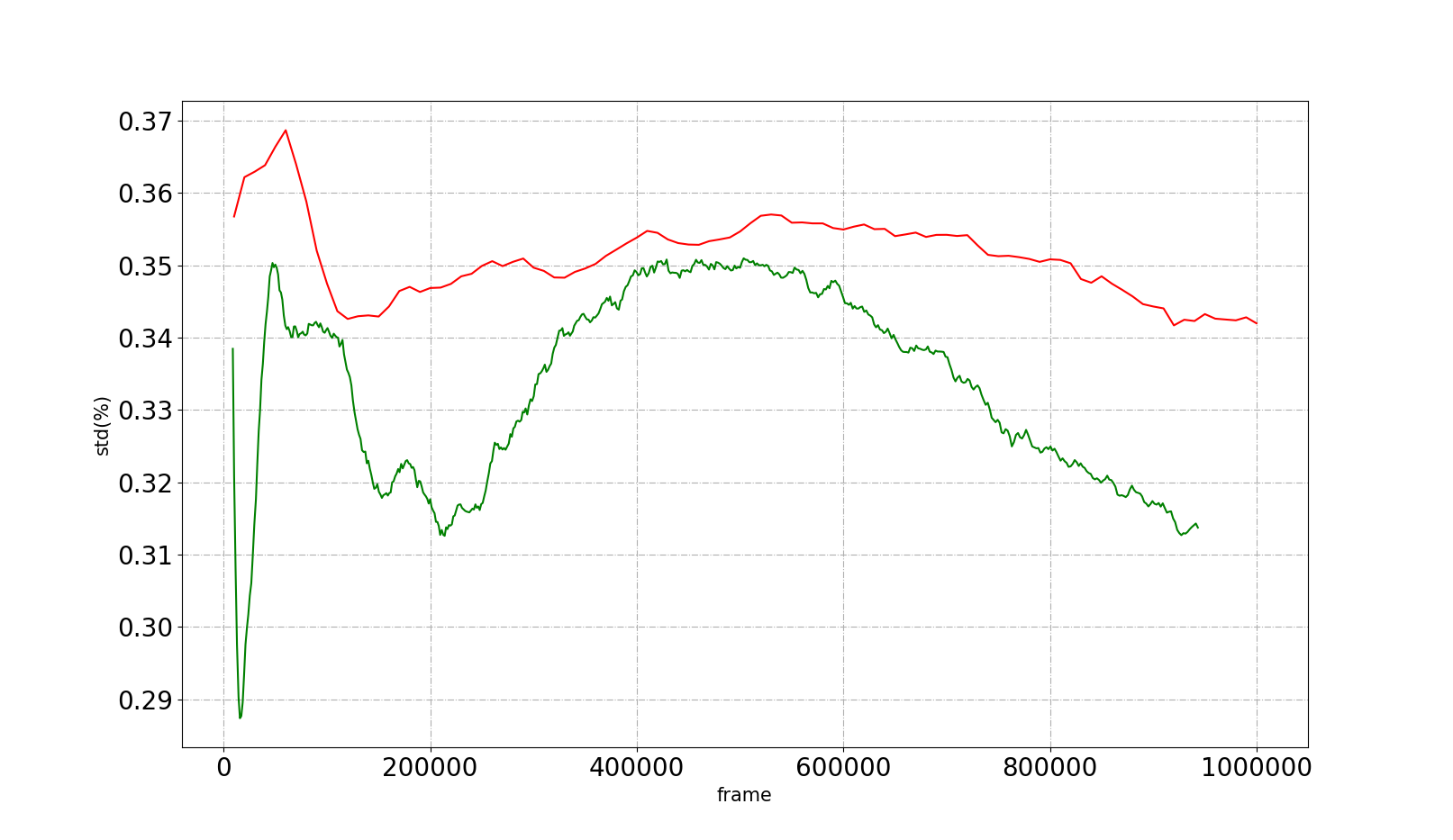}
			\end{minipage}%
		}%
		\centering
		\caption{Standard deviation curves during learning process}	
		\label{fig:std}
	\end{figure*}

	The idea of Theorem \ref{thm3} is inspired by Deep Double $Q$-learning \cite{van2016deep}. We first assume that the $Q$ function is unbiased, and then prove that the $Q$ function will always be biased after learning with the target function. Theorem \ref{thm3} shows that even if it is assumed that the $Q$ function in TD3 is unbiased after learning, the update process in TD3 always leads to a downward deviation of the $y$ estimation, and the size of this deviation is related to the variance of the $Q$ value. Therefore, we can believe that when using the Clipped Double Q-learning method to update the target value, if the variance of the $Q$ value of an algorithm is smaller, then this algorithm can alleviate the downward deviation that Clipped Double Q-learning strategy brings. We tested the standard deviations of the $Q$ value of RUD and TD3 during the learning process in Hopper and Walker2d. Each time 1000 states were sampled from the replay buffer to calculate the $Q$ value and standard deviation. We only showed the change in the standard deviation of $Q_1$. This is because $Q_1$ and $Q_2$ have obvious similarities in the learning process \cite{fujimoto2018addressing}. As shown in Figure \ref{fig:std}, especially in the late learning period, the standard deviation of the $Q$ value of RUD is lower than TD3.
	
	We believe that, compared with the policy that is updated at every time step in TD3, the policy of RUD for exploration remains unchanged during every $F$ time steps, which leads to the diversity of state-action pairs in the RUD replay buffer being less than TD3. As a result, the standard deviation of the $Q$ value of RUD during the learning process can be significantly lower than that of TD3. Insufficient variety of state-action pairs may lead to insufficient exploration, but we can still adjust the $F$ parameter of RUD to obtain the balance between exploration and exploitation. Overall, RUD's low standard deviation of Q value makes it more suitable for using Clipped Double Q-learning strategy than TD3.

	\section{Experiments}
	The experiments of this section will answer the following questions: 
	
	(1) Can RUD perform better than previous algorithms on the Mujoco environments? 
	
	(2)  Without introducing other optimization tricks, how much improvement can the Regularly Updated reinforcement learning paradigm bring comparing to the original DDPG method? 
	
	(3) How does the performance of RUD change with the change of the parameter $F$? 
	
	(4) Does the characteristic that RUD can replay new experiences with greater probability improve its data utilization?
	
	We use continuous control tasks provided by Mujoco \cite{todorov2012mujoco} based on OpenAI Gym \cite{dhariwal2016openai} to evaluate performance of the algorithms. These tasks are widely used in the evaluation of reinforcement learning in continuous environments \cite{schulman2017proximal} \cite{wu2017scalable} \cite{haarnoja2018soft}. In order to further ensure the fairness of the test results, all test results are averaged over 10 times under at least 5 different random seeds. The code of RUD is implemented based on Pytorch \cite{adam2017automatic}. In the comparative experiment, some of the previous algorithms are implemented using Tensorflow \cite{abadi2016tensorflow}.
	
	\subsection{Comparison}
	We first compare the performance of RUD with the previous algorithms on the Mujoco environments to confirm the superiority of RUD. We compared the performance of RUD with DDPG \cite{lillicrap2015continuous}, Proximal Policy Optimization (PPO) \cite{schulman2017proximal}, and Twin Delayed Deep Deterministic (TD3) policy gradient algorithm \cite{fujimoto2018addressing} in 6 environments. DDPG and PPO are implemented using the code provided by OpenAI Baselines \cite{dhariwal2016openai}. The parameters of these algorithms are set to the default optimal parameters in Baselines. TD3 is implemented using the code provided by its authors Scott Fujimoto et al. \footnote{The code is available on the website: https://github.com/sfujim/TD3}. and the parameter setting is consistent with the original paper \cite{fujimoto2018addressing}.
	
	We use feedforward neural networks with two hidden layers to implement the actor and critic networks in RUD. Linear rectifier units (ReLU) are used between hidden layers. The last layer of the actor network is linked to tanh units. The two hidden layers contain 400 and 300 neurons respectively. Both actor and critic networks learn using Adam optimizer \cite{kingma2014adam}. A mini-batch with 256 transition pairs is sampled from the replay buffer for the network training, the learning rate is set to $ 0.3 \times10 ^ {-4} $.
	
	RUD applies the Target Policy Smoothing and Delayed Policy Update techniques of TD3. The output of the target policy is added with Gaussian noise $\mathcal N\sim(0,0.2)$, and the noise is clipped to the range $[-0.5, 0.5]$. Actor networks and the target critic network are updated every two iterations. The soft update weight $\tau$ of all target networks is set to $0.005 $. The Gaussian noise for exploration is $ \mathcal N \sim (0, 0.1) $. The fixed frame number $F$ to maintain the learned policy for exploration without changing is set to 2500 in Halfcheetah and Walker-2d, 1250 in the other tasks. The detailed performance of RUD using different $F$ parameters in different environments can be found in Section 6.3.
	
\begin{figure*}[tb]	
		\centering
		\subfigure[Ant-v2]{
			\begin{minipage}[t]{0.33\linewidth}
				\centering
				\includegraphics[width=1.5in]{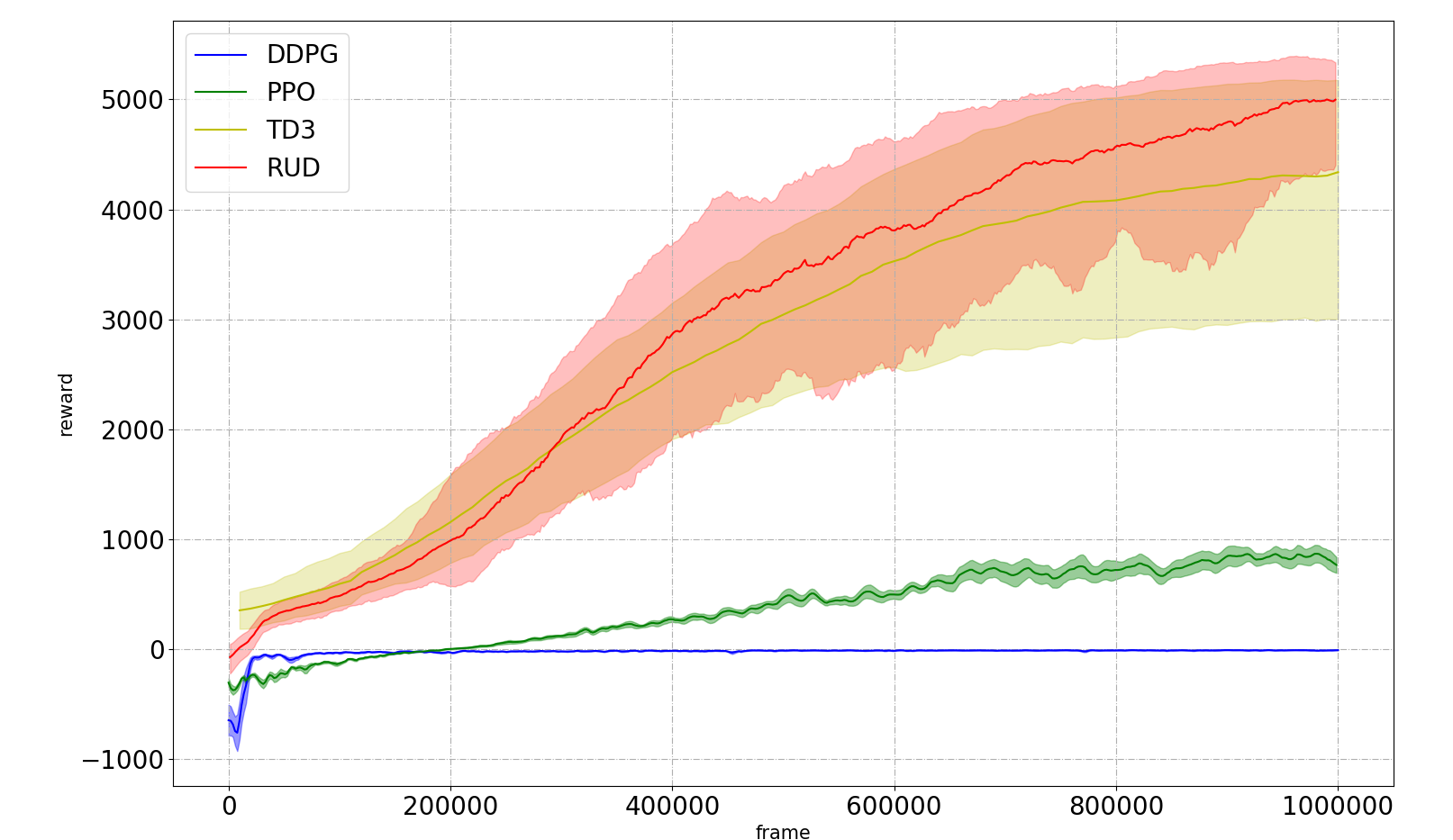}
			\end{minipage}%
		}%
		\subfigure[Hopper-v2]{
			\begin{minipage}[t]{0.33\linewidth}
				\centering
				\includegraphics[width=1.5in]{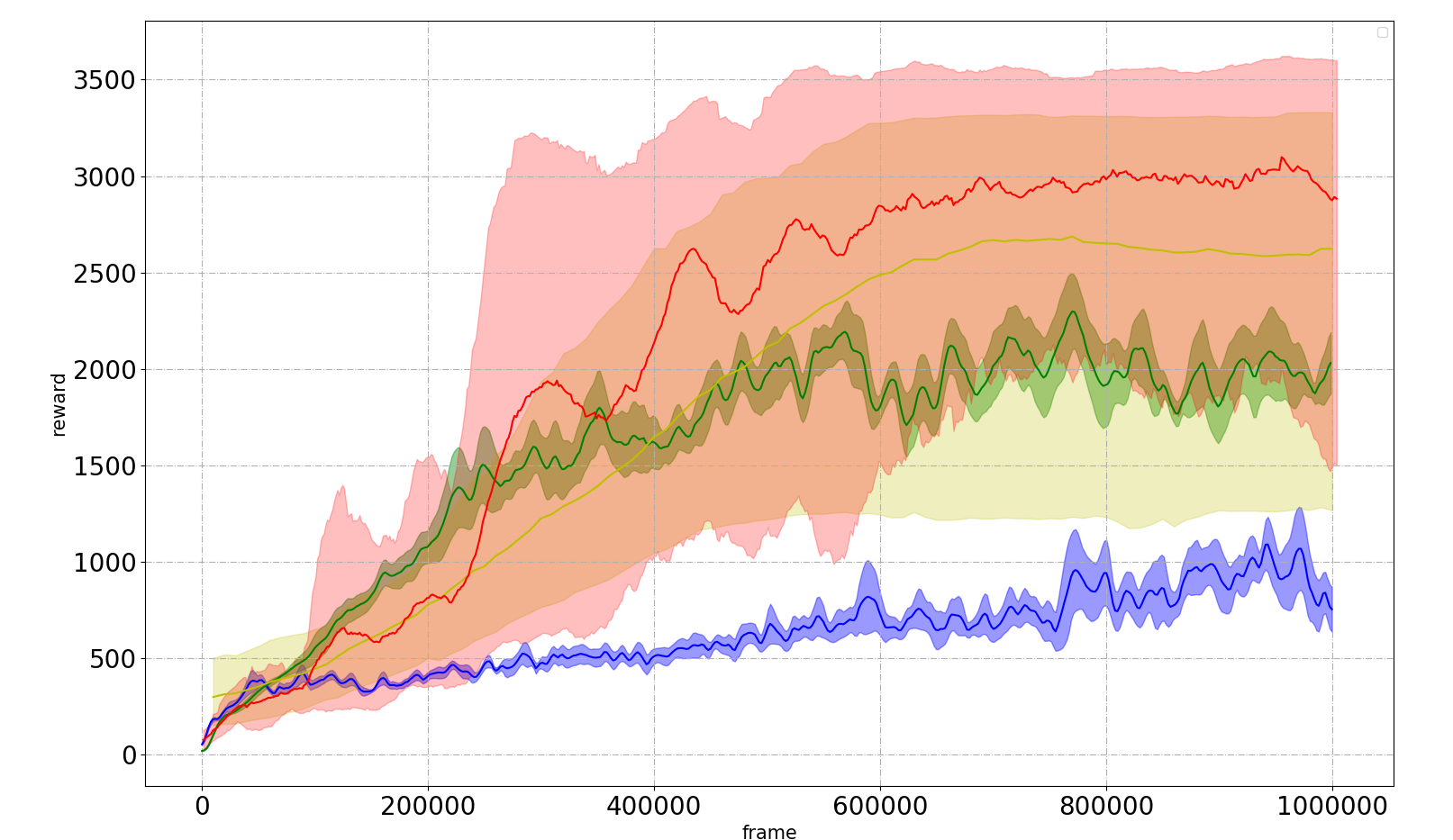}
			\end{minipage}%
		}%
		\subfigure[HalfCheetah-v2]{
			\begin{minipage}[t]{0.33\linewidth}
				\centering
				\includegraphics[width=1.5in]{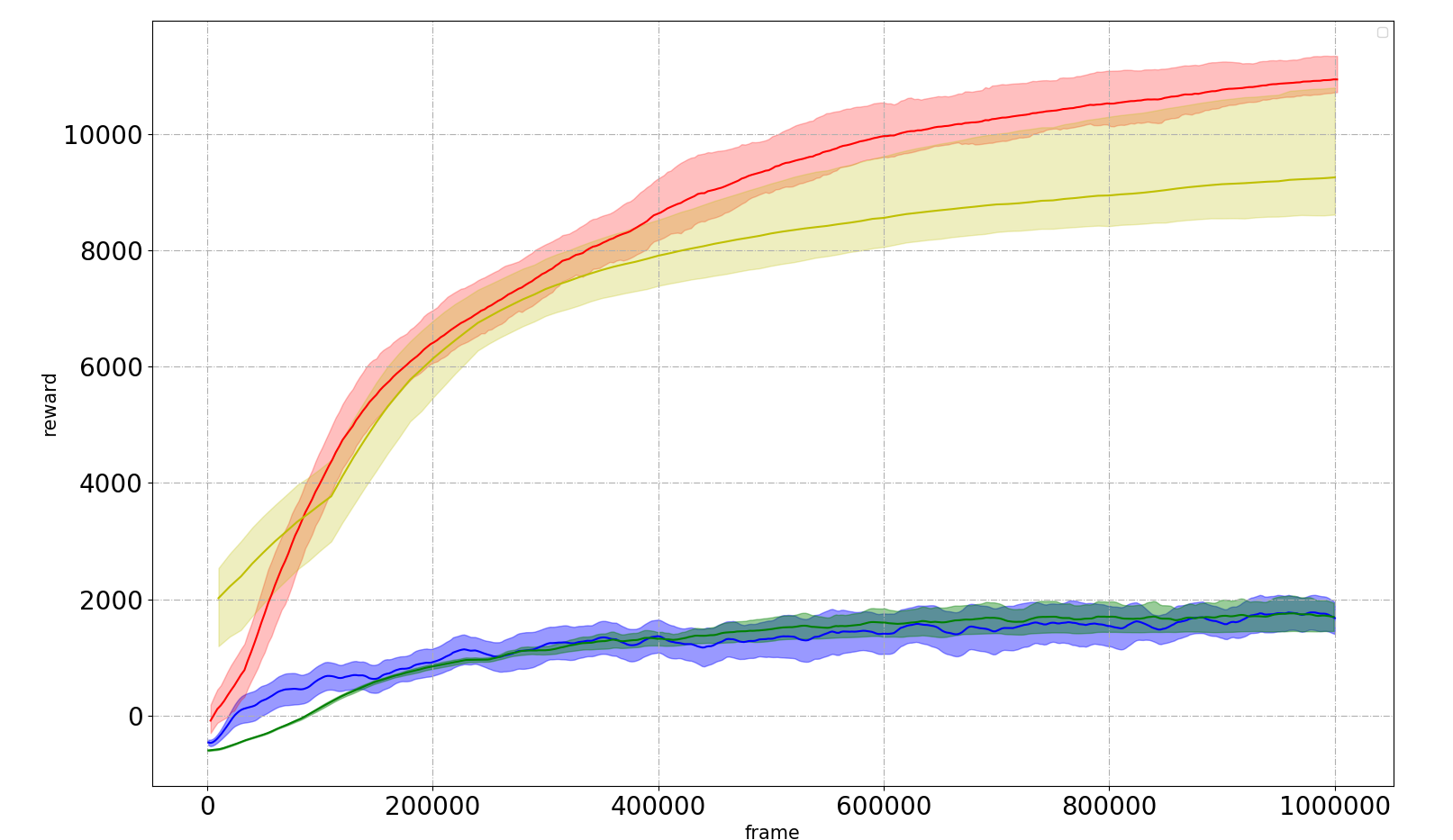}
			\end{minipage}%
		}%
		
		\subfigure[Walker2d-v2]{
			\begin{minipage}[t]{0.33\linewidth}
				\centering
				\includegraphics[width=1.5in]{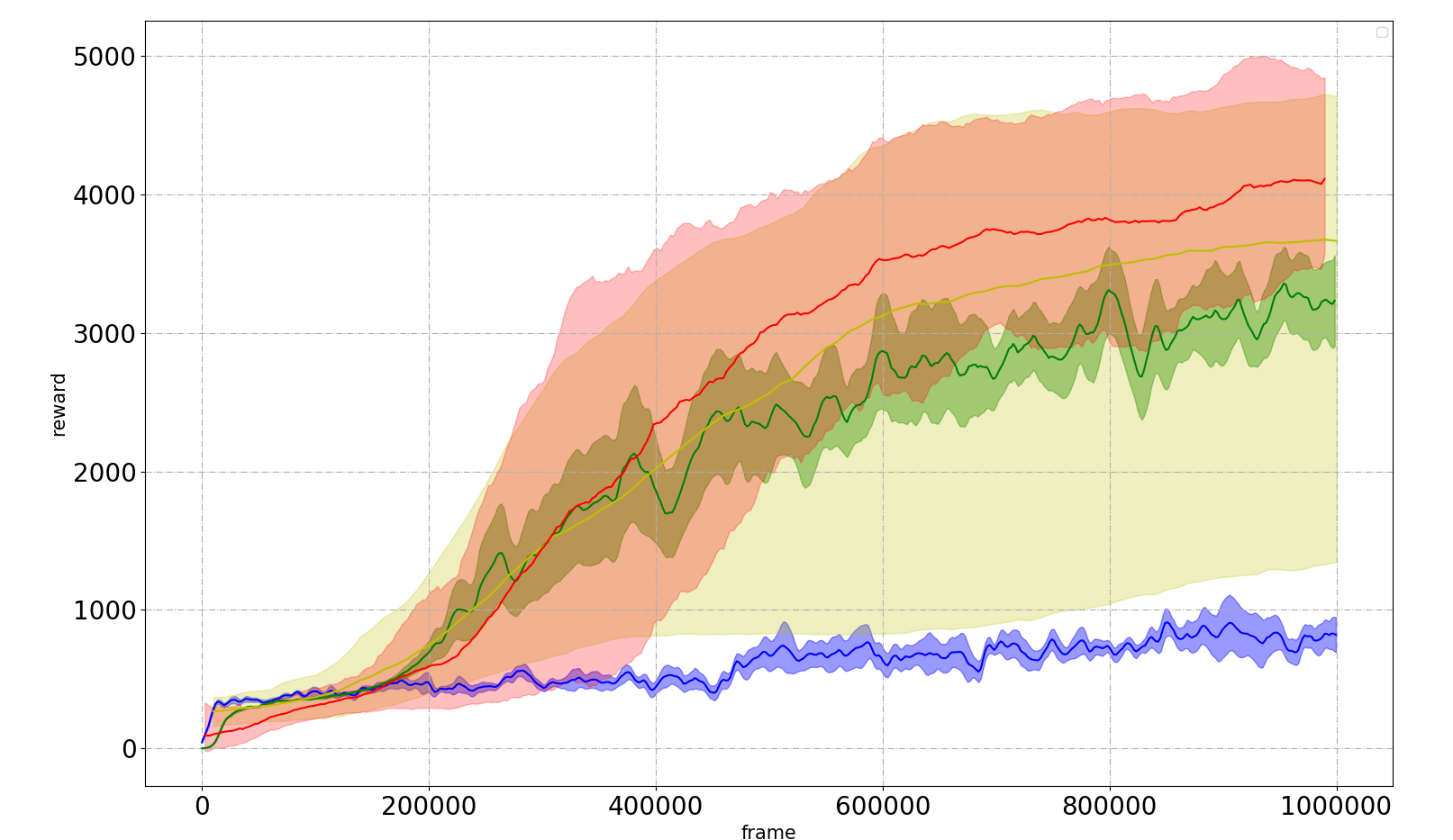}
			\end{minipage}%
		}%
		\subfigure[InvertedDoublePendulum-v2]{
			\begin{minipage}[t]{0.33\linewidth}
				\centering
				\includegraphics[width=1.5in]{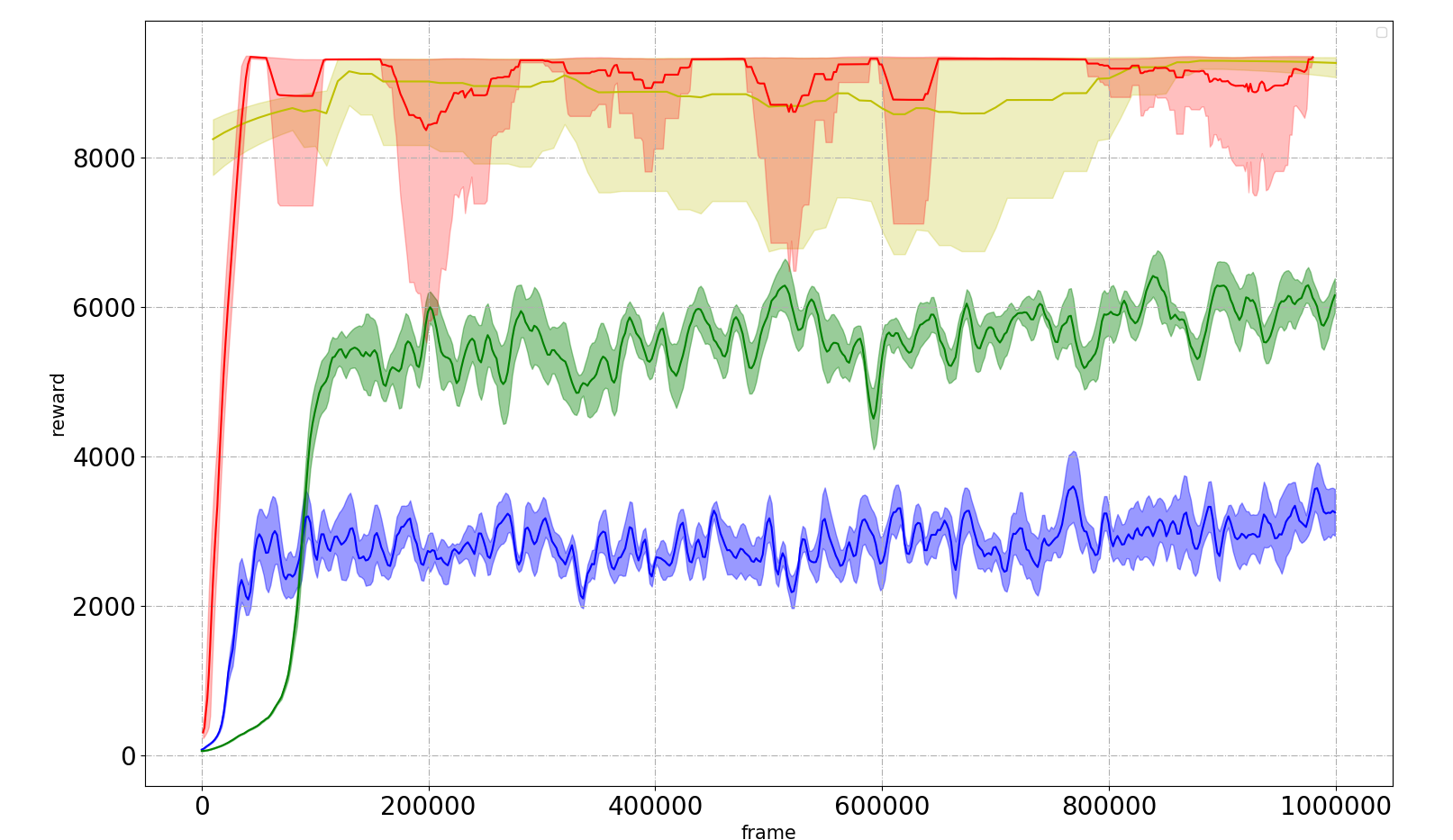}
			\end{minipage}%
		}%
		\subfigure[InvertedPendulum-v2]{
			\begin{minipage}[t]{0.33\linewidth}
				\centering
				\includegraphics[width=1.5in]{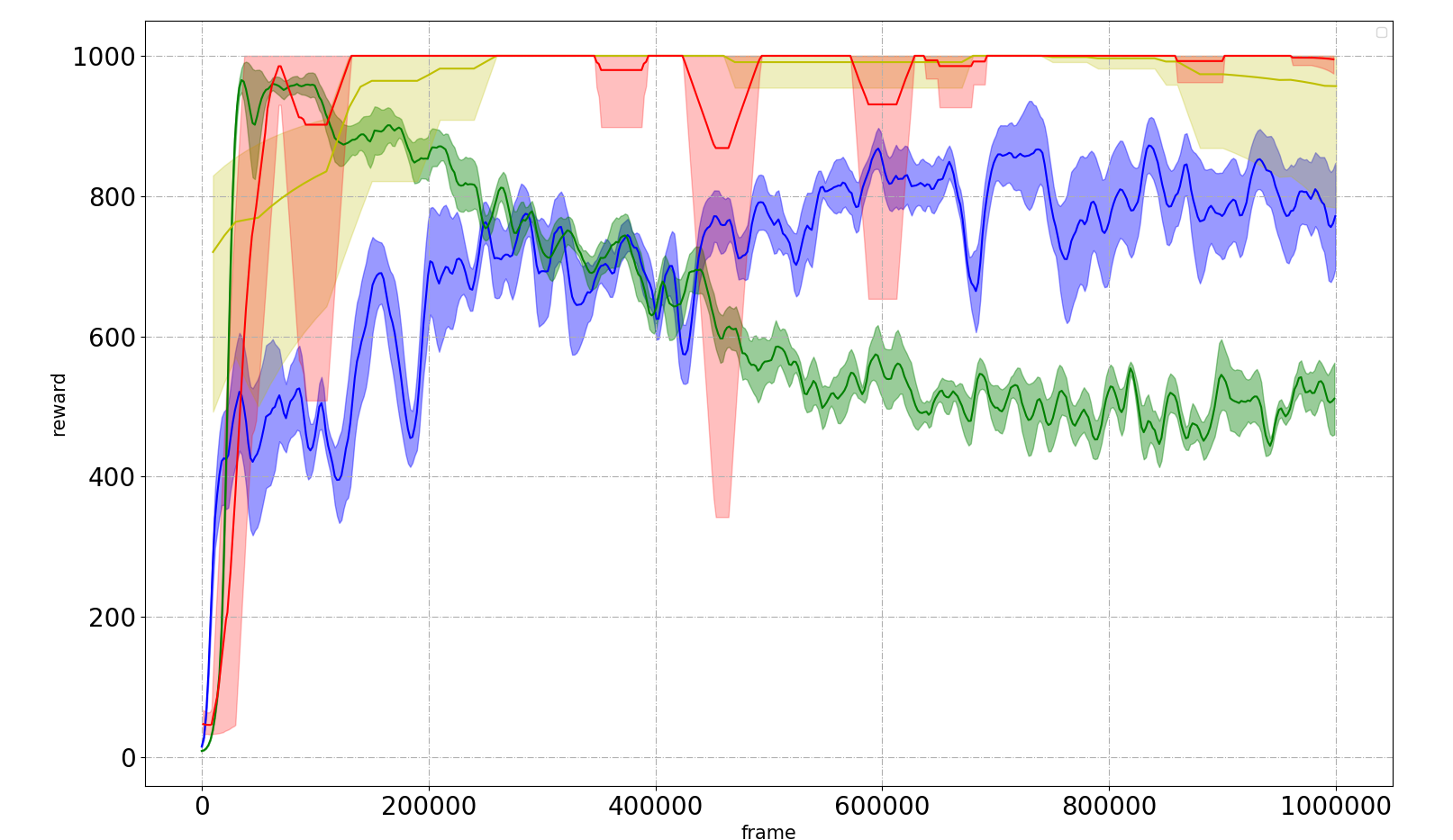}
			\end{minipage}%
		}%
		
		\centering
		\caption{Learning curves on Mujoco-based continous control benchmarks.}	
		\label{fig:compare_4}
	\end{figure*}

	Each task performed 1 million time steps, algorithms were evaluated every $F$ steps. The noise used for exploration is removed during evaluation. And each evaluation adopt the average reward over 10 test results. The evaluation of each algorithm uses at least 5 sets of random seeds to initialize networks and Gym simulator. The results are shown in Figure \ref{fig:compare_4}. Compared with the previous method, the stability of RUD in Ant, Hopper, Halfcheetah, and Walker2d is significantly better than that of TD3, and the average performance is significantly better than all other methods. This is because these environments involve more complex state-action spaces. RUD's higher data utilization makes its performance significantly better than previous methods in these more challenging environments. InvertedDoublePendulum and InvertedPendulum are relatively uncomplicated, so both TD3 and RUD can achieve good performance in these environments.

	\subsection{Ablation Experiment}
	
	TD3 uses a lot of optimization techniques for DDPG. While these techniques bring a huge improvement to the original DDPG, they also mask the improvement made by RUD to the original DDPG. To further confirm the effectiveness of the regularly updated paradigm adopted by RUD, we evaluated the performance of DDPG adopting the regularly updated paradigm without any other optimization skills and compared it with the performance of original DDPG.
	
\begin{figure*}[!h]	
		\centering
		\subfigure[Hopper-v2]{
			\begin{minipage}[t]{0.33\linewidth}
				\centering
				\includegraphics[width=1.5in]{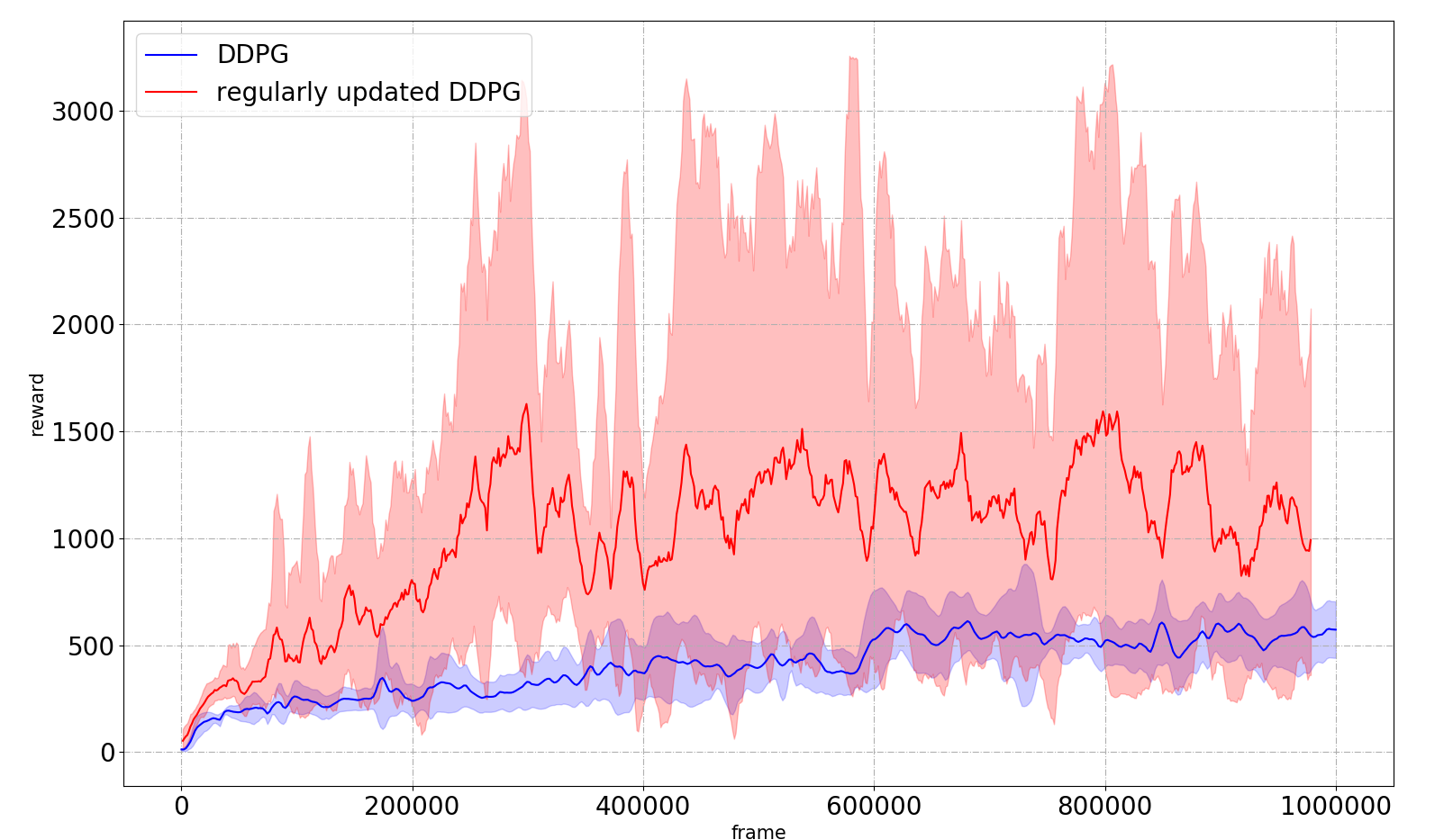}
			\end{minipage}%
		}%
		\subfigure[Ant-v2]{
			\begin{minipage}[t]{0.33\linewidth}
				\centering
				\includegraphics[width=1.5in]{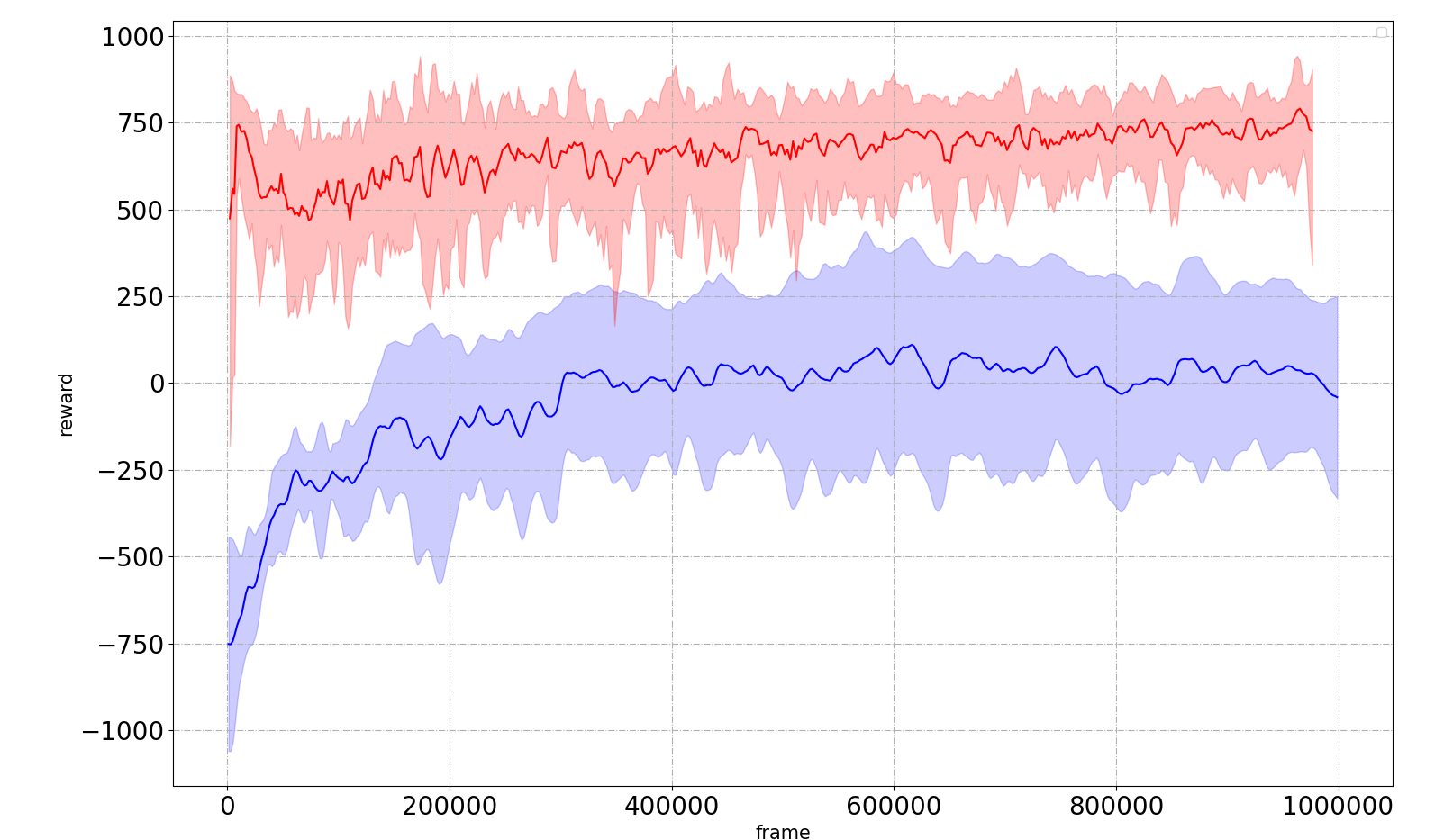}
			\end{minipage}%
		}%
		\subfigure[HalfCheetah-v2]{
			\begin{minipage}[t]{0.33\linewidth}
				\centering
				\includegraphics[width=1.5in]{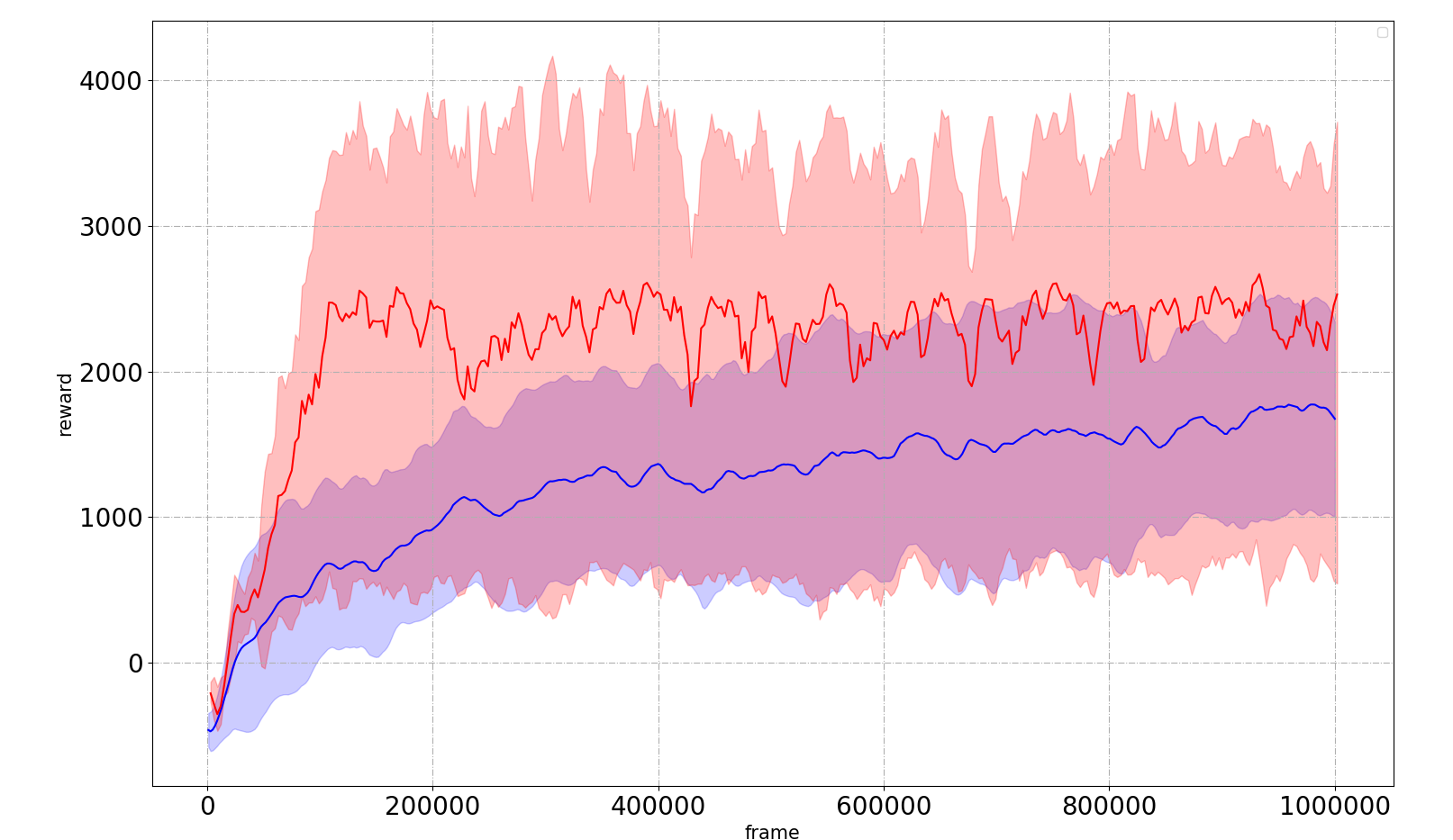}
			\end{minipage}%
		}%
		
		\subfigure[Walker2d-v2]{
			\begin{minipage}[t]{0.33\linewidth}
				\centering
				\includegraphics[width=1.5in]{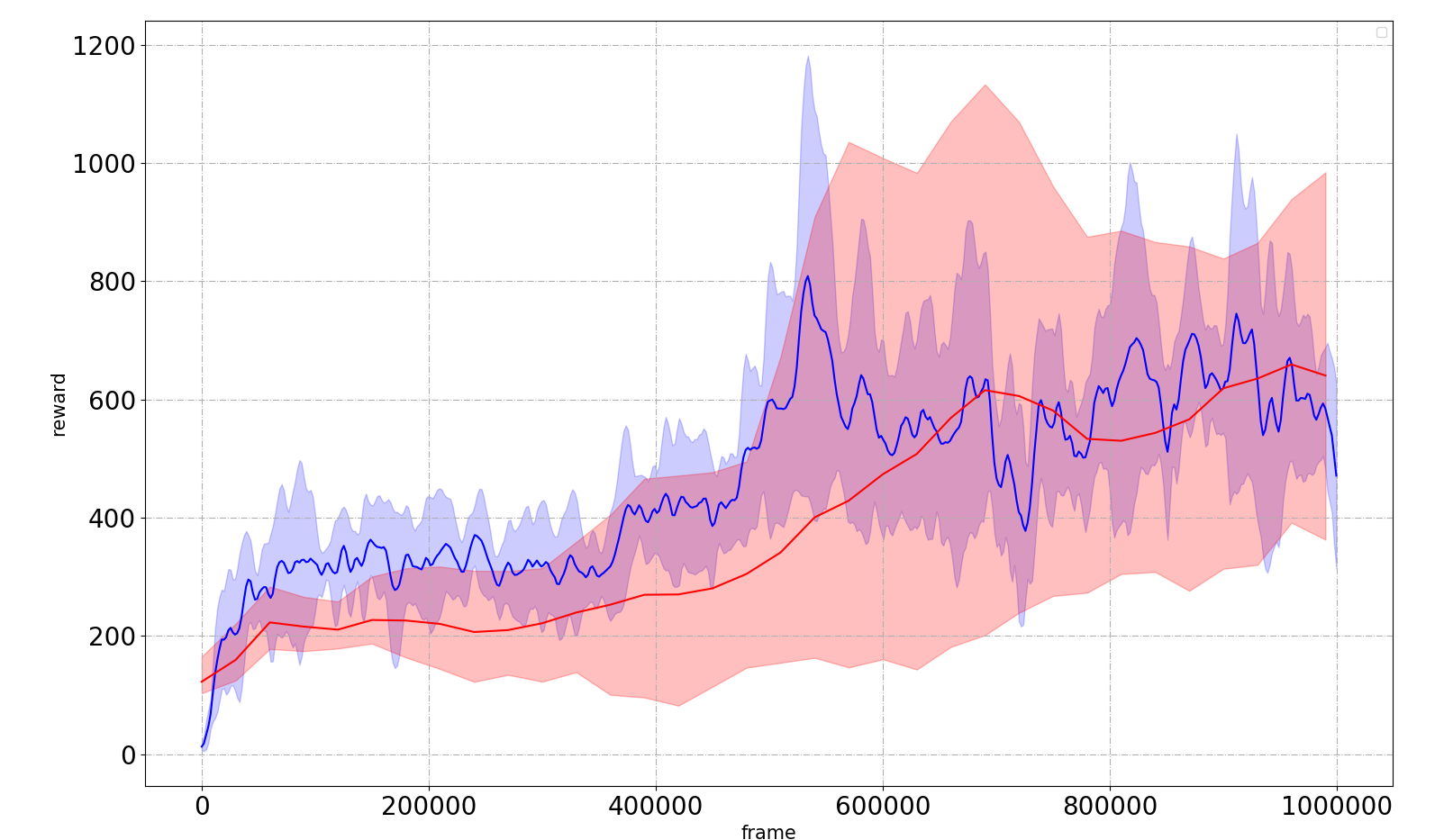}
			\end{minipage}%
		}%
		\subfigure[InvertedDoublePendulum-v2]{
			\begin{minipage}[t]{0.33\linewidth}
				\centering
				\includegraphics[width=1.5in]{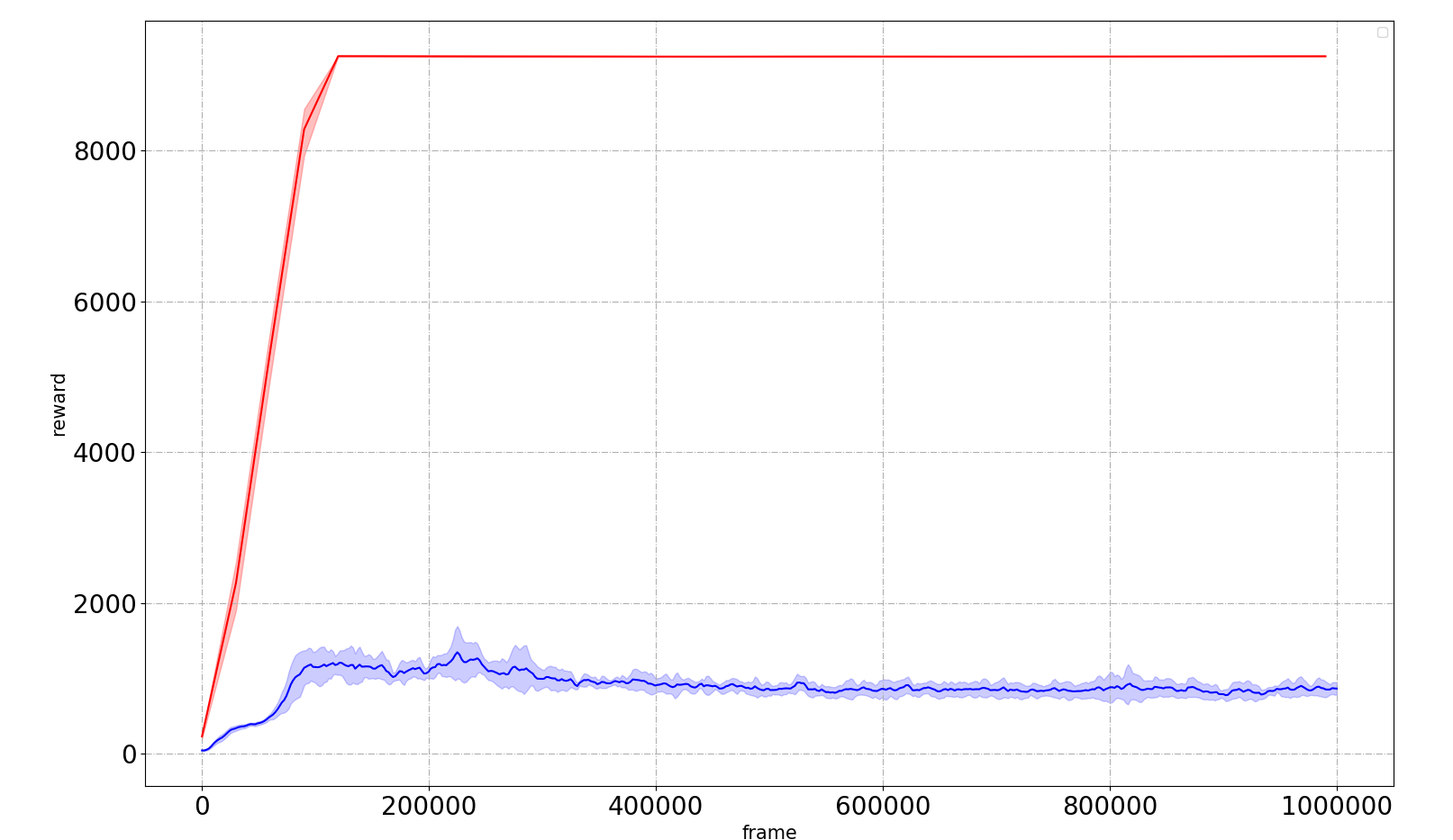}
			\end{minipage}%
		}%

		\centering
		\caption{Comparison of Regularly Updated DDPG and original DDPG.}	
		\label{fig:compare_2}
	\end{figure*}

	The parameter setting of the DDPG in this subsection is consistent with that in the original paper \cite{lillicrap2015continuous}, except that the batch-size is set to $128$, which has been shown to improve performance in recent work \cite{islam2017reproducibility}. The off-policy exploration noise is the Ornstein-Uhlenbeck process. The fixed exploration frame number $F$ of DDPG with the regularly updated paradigm is set to $1250$. The results are shown in Figure \ref{fig:compare_2}.	 Except in the Walker2d environment where the effect is not obvious, the regularly updated DDPG performs significantly better than the original DDPG in other environments. In the context of Ant and InvertedDoublePendulum, the worst performance of regularly updated DDPG are significantly better than the best performance of the original DDPG.

	\subsection{Setting of Parameter F}
	
	An important issue in the RUD is how to select the parameter $F$. Intuitively, if $F$ is larger, the agent will use the data in the replay buffer more fairly, but it also means that the agent's exploration efficiency is lower. An extreme example is when $F$ is set to 1 million that is equal to the total time step of the task execution. Under this circumstance, when the agent is sampling from the replay buffer, each sampling is to uniformly sample from the all 1 million experiences, and the probability of each experience being sampled is equal. However, it also means that the exploration policy of the entire interaction process with the environment is only a disturbance to the initial policy. Thence the exploration efficiency may be very low. Another extreme example is when $F$ is set to 1, then RUD will be consistent with the classical reinforcement learning algorithm. Each time step the agent will interact with the environment and learn according to samples from the current replay buffer. According to Theorem \ref{thm1}, the agent is underutilized for new experience, but at the same time, because the actor network (i.e. policy) is updated at each time step, the exploration policy at each time step is also different. Consequently, the exploration efficiency of the agent is the high. Therefore, the setting of parameter $F$ is essentially a balance between exploitation and exploration.
	
\begin{figure*}[!htb]
		\centering
		\subfigure[Ant-v2]{
			\begin{minipage}[t]{0.5\linewidth}
				\centering
				\includegraphics[width=2.5in]{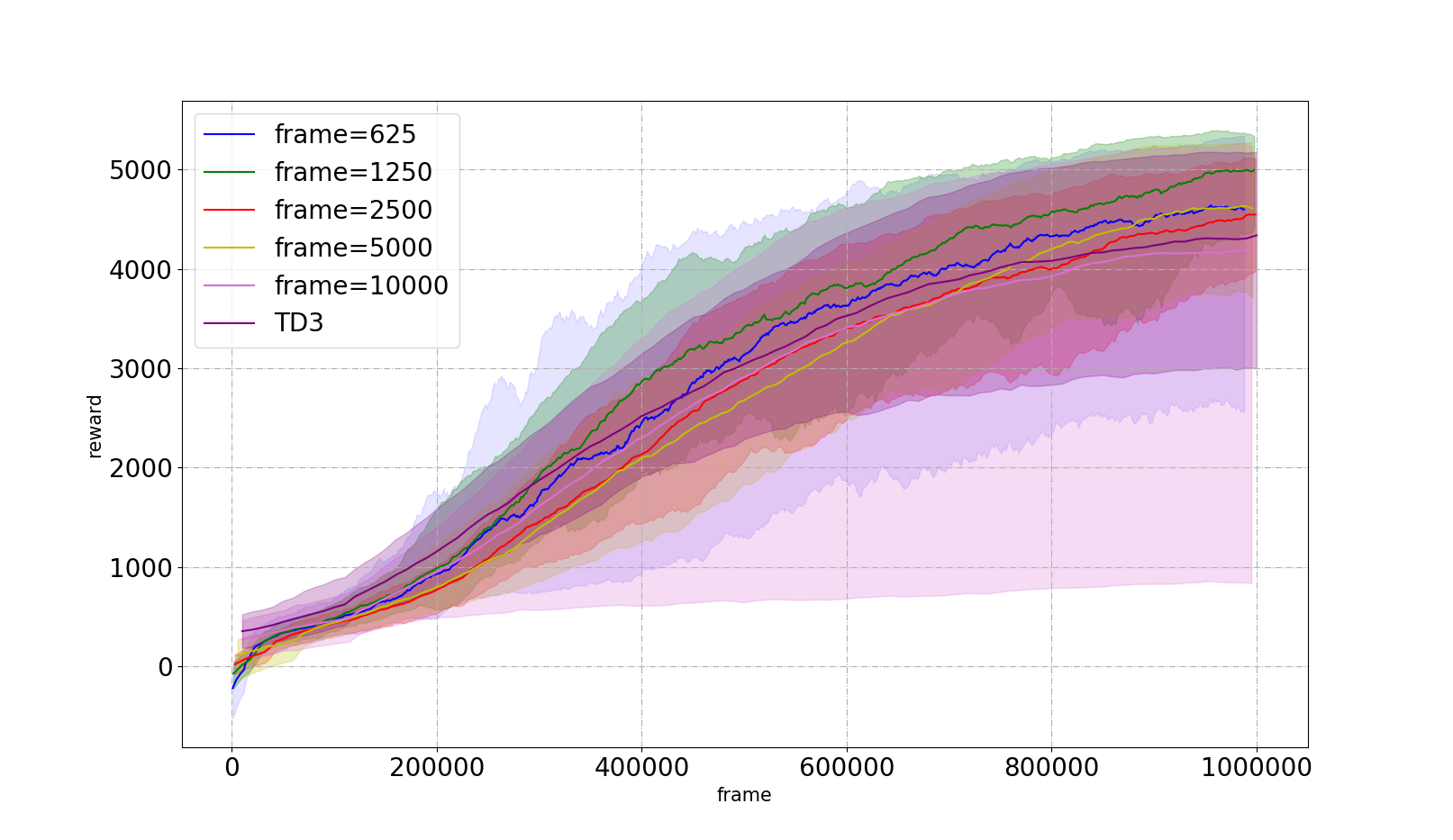}
			\end{minipage}%
		}%
		\subfigure[Halfcheetah-v2]{
			\begin{minipage}[t]{0.5\linewidth}
				\centering
				\includegraphics[width=2.5in]{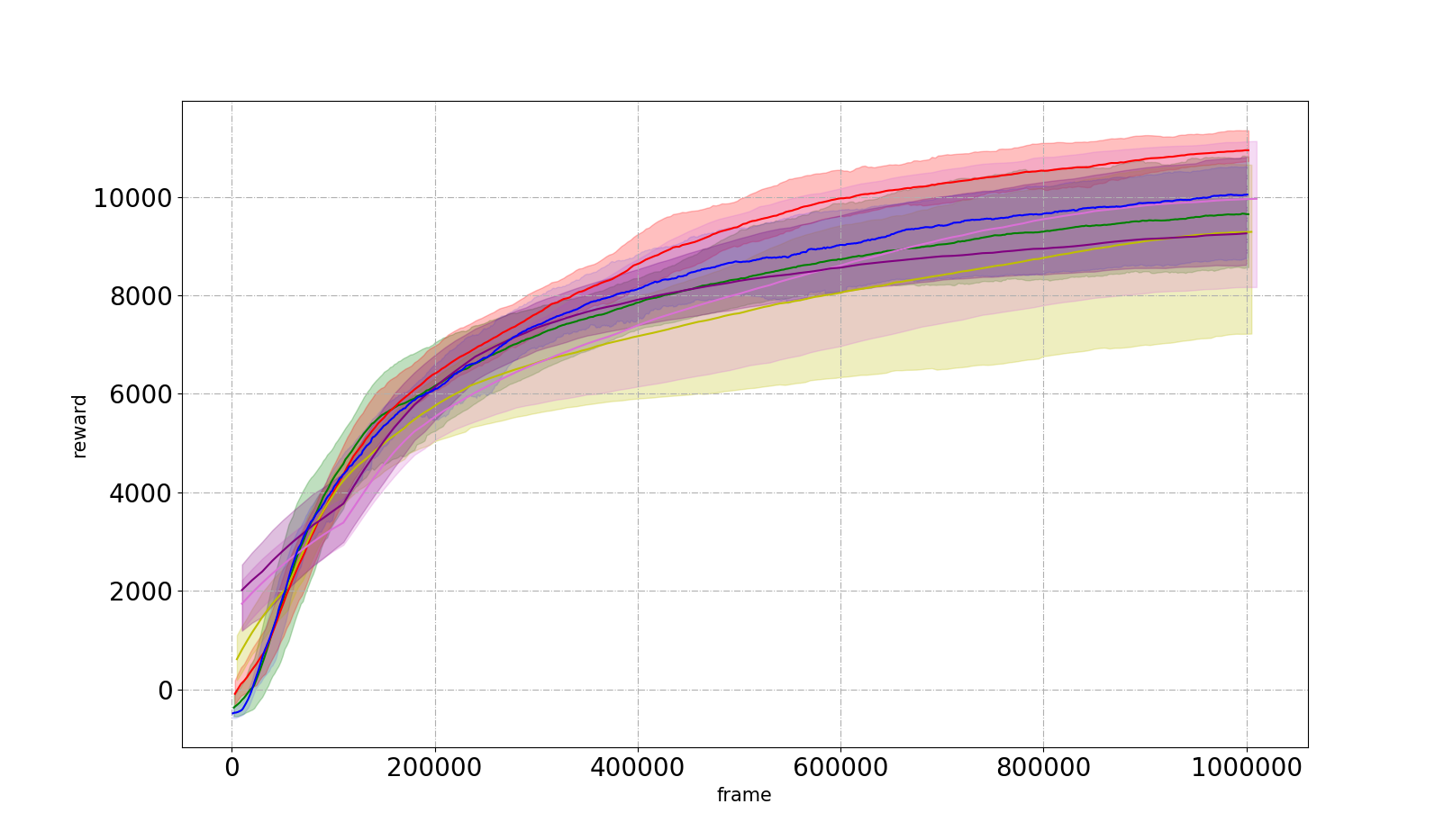}
			\end{minipage}%
		}%
		
		\subfigure[Walker2d-v2]{
			\begin{minipage}[t]{0.5\linewidth}
				\centering
				\includegraphics[width=2.5in]{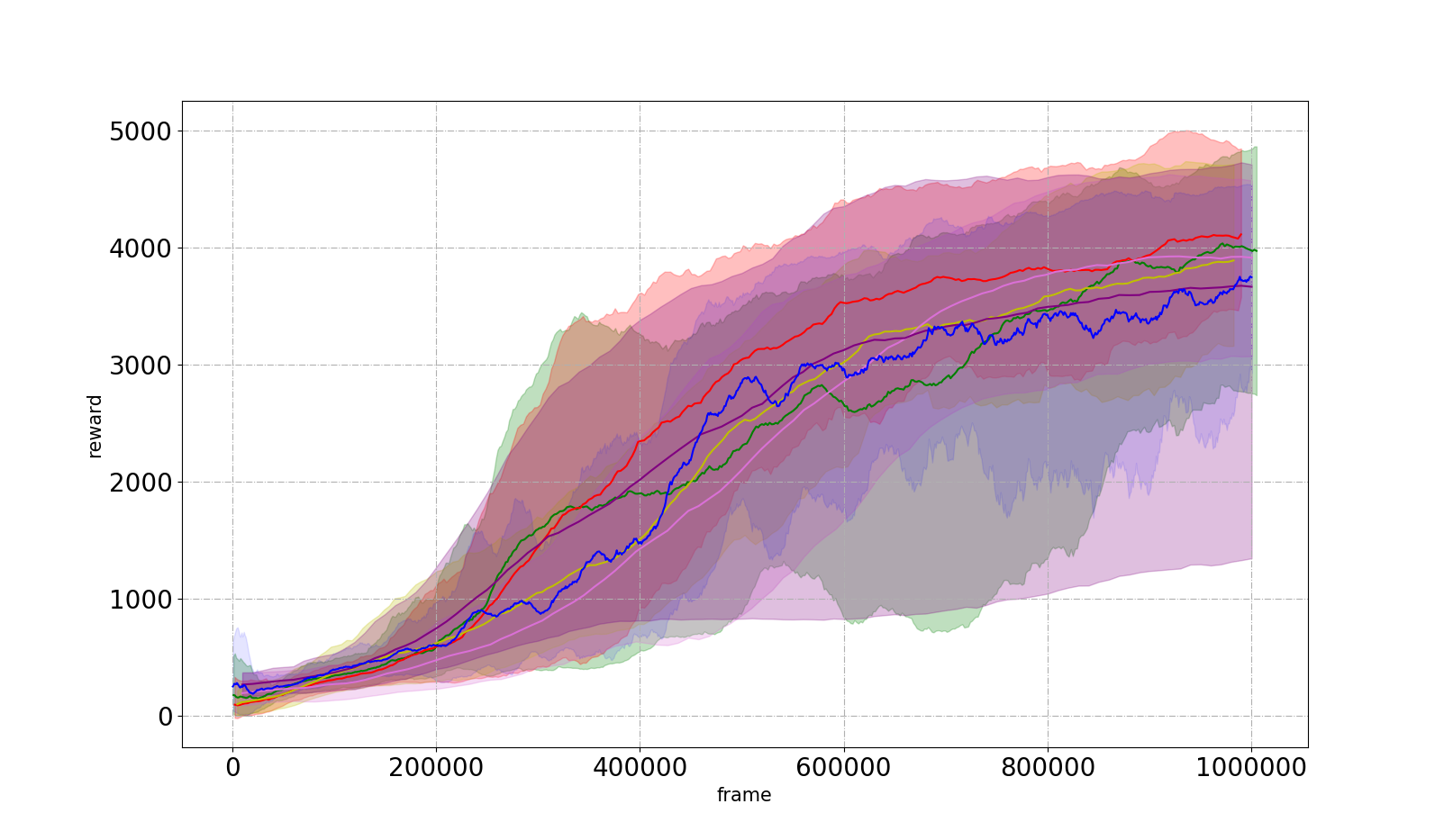}
			\end{minipage}%
		}%
		\subfigure[Hopper-v2]{
			\begin{minipage}[t]{0.5\linewidth}
				\centering
				\includegraphics[width=2.5in]{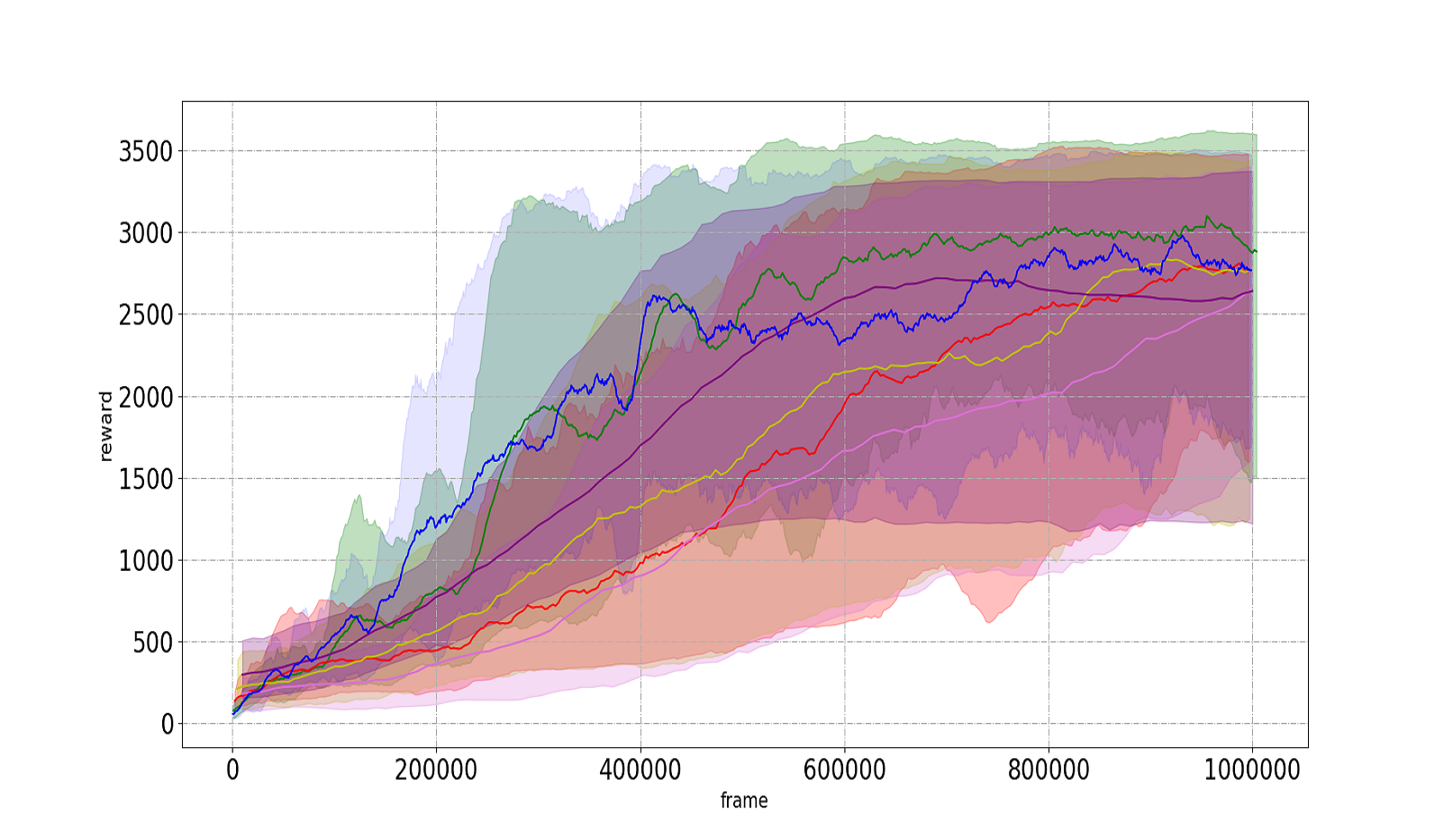}
			\end{minipage}%
		}%
		\centering
		\caption{Performance of RUD under different parameter settings.}	
		\label{fig:F}
	\end{figure*}

	In this subsection, we will test how the performance of RUD changes with the different parameter $F$ in four environments. We will take the fixed frame $F$ from $\{10000, 5000, 2500, 1250, 625\}$ to test the performance of RUD, and compare them with TD3 (i.e. $F=1$). The experimental results are shown in Figure \ref{fig:F}.  Results show that when $F$ is too large or too small, the performance of RUD is not very good. In HalfCheetah and Ant environments, RUD is better than TD3 in most cases. RUD works best when $F$ is set to 2500 in HalfCheetah and Walker2d, and works best when $F$ is set to 1250 in Hopper and Ant.

	\subsection{Learning Efficiency of RUD}
	
	The distinguishing feature of RUD from classical reinforcement learning methods is that RUD can replay new experiences with greater probability. This subsection will explore whether RUD's ability to replay new experiences with greater probability improves its data utilization. For the same action $a_t$ and state $s_t$, we will take the $\triangledown = |Q(s_t, a_t)-Q'(s_t, a_t)|$ to judging the effect of an update, where $Q(s_t, a_t)$ is the $Q$ value before this update and $Q'(s_t, a_t)$ is the $Q$ value after this update. The larger $\triangledown$ is, the more effective the update is, otherwise the update is more invalid.
	
\begin{figure*}[!th]	
		\centering
		\subfigure[Walker2d-v2]{
			\begin{minipage}[t]{0.5\linewidth}
				\centering
				\includegraphics[width=2.5in]{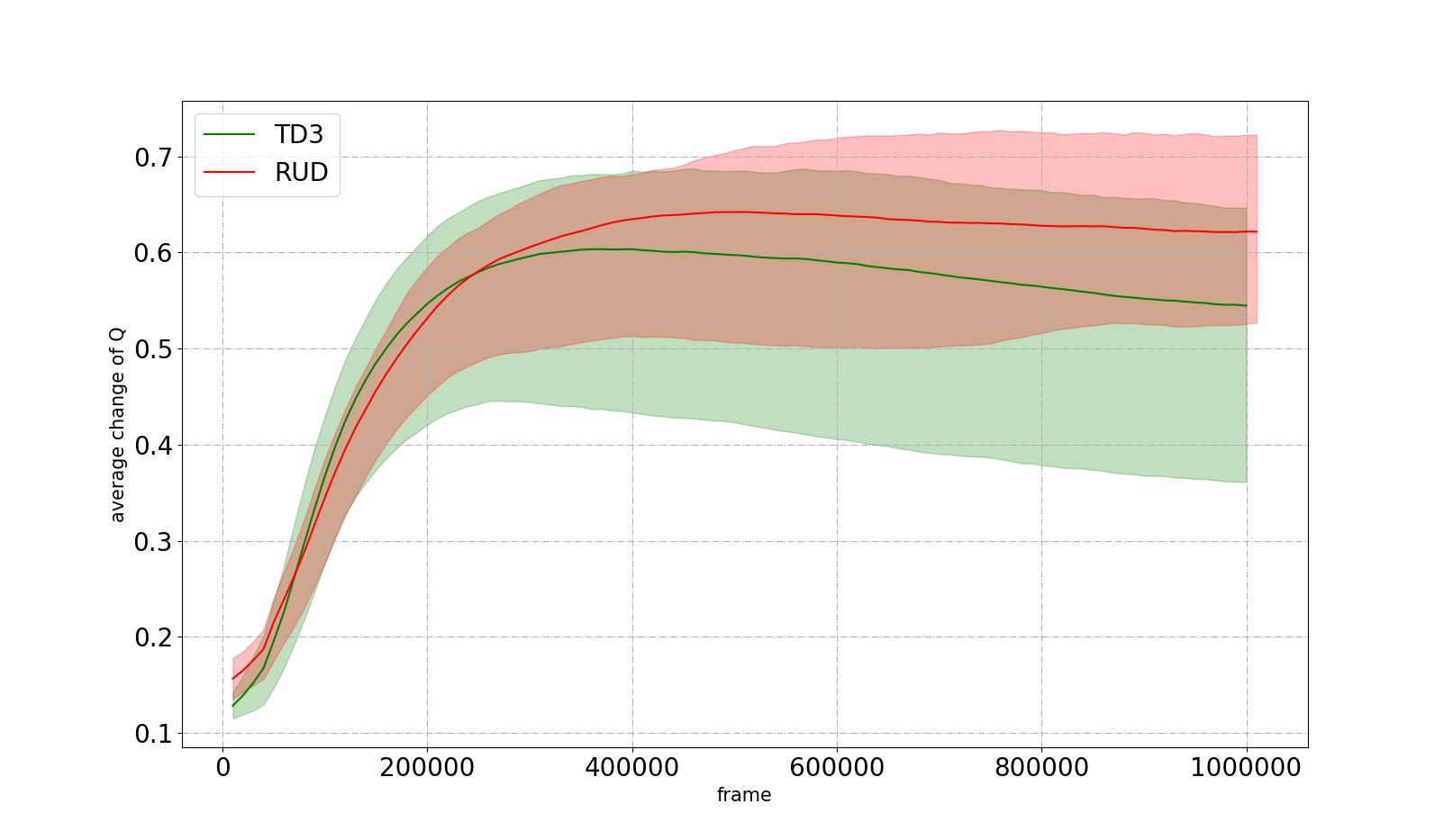}
			\end{minipage}%
		}%
		\subfigure[Hopper-v2]{
			\begin{minipage}[t]{0.5\linewidth}
				\centering
				\includegraphics[width=2.5in]{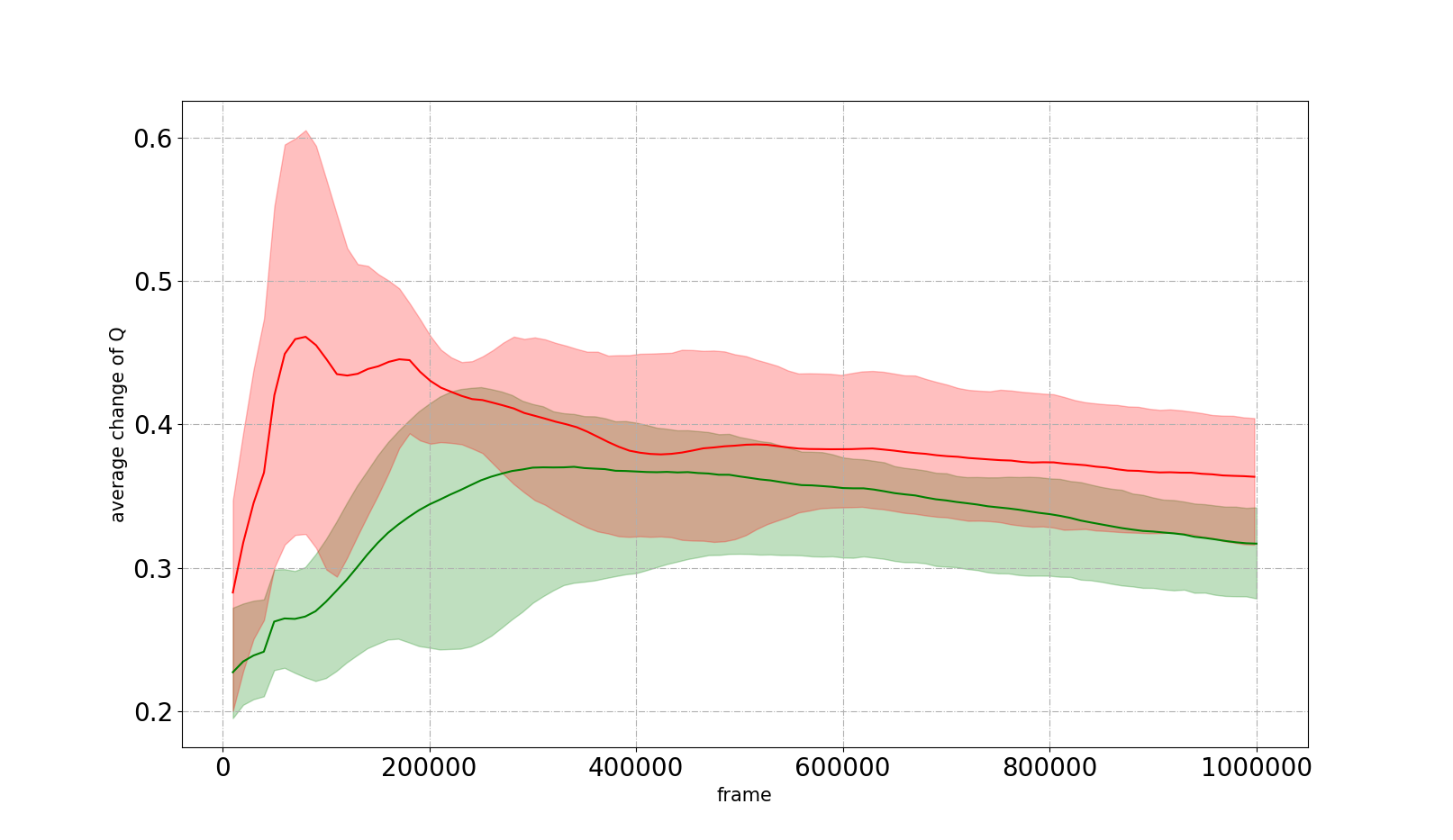}
			\end{minipage}%
		}%
		\centering
		\caption{Change of Q value on Walker2d-v2 and Hopper-v2}	
		\label{fig:changeofQ}
	\end{figure*}

	When evaluating the update effect of the algorithm, we uniformly sample 1000 state-action pairs $(s_i, a_i), i=1, 2, ..., 1000$ in the replay buffer, then we calculate the sum of $\triangledown_{i}$ and take the average: $\frac{1}{1000} \sum_{i=1}^{1000}\triangledown_{i} = \frac{1}{1000} \sum_{i=1}^{1000} |Q '(s_i, a_i)-Q(s_i, a_i)|$. Both RUD and TD3 are evaluated and recorded every $F$ frames. The results are shown in Figure \ref{fig:changeofQ}.	 In Walker2d and Hopper, the average $Q$ change of TD3 starts to decline at around 300,000 frames, because in the subsequent update process after 300,000 frames, TD3 still samples more old experiences, which have already been learned. Although the average $Q$ change of RUD also begins to decline at around 300,000 frames, the decline is significantly smaller than TD3, which shows that RUD uses more new experience than TD3 and these experiences has rarely been used before, so the data utilization of RUD is higher than TD3.

\section{Conclusion and Future Work}
In order to solve the problem of poor data utilization of the classical DPG algorithms, we proposed the Regularly Updated Deterministic policy gradient method. The main contributions of this paper are as follows:

(1) We theoretically analyzed the data utilization problem of the classical DPG methods and pointed out that under the classical DPG procedure, the agent always overuses the old experiences and rarely uses the new experiences.

(2) We proposed the Regularly Updated Deterministic policy gradient method, and theoretically proved that RUD can better use new experiences than classical DPG.

(3) We theoretically pointed out that the Clipped Double Q-learning strategy will bring deviation to the target, and empirically show that RUD can effectively alleviate the deviation brought by the Clipped Double Q-learning strategy.

In the experimental part of this paper, the superiority of the Regularly Updated paradigm in the DPG method is verified by the comparison experiment. And we further confirmed the superiority of RUD through ablation experiment, parameter exploration experiment and data utilization analysis experiment. 

In future work, it can be investigated whether the Regularly Updated paradigm still has superiority in Deep Q-learning \cite{mnih2015human}, Soft Actor-critic \cite{haarnoja2018soft}, and other off-policy methods.

\section*{Acknowledgement}

This work was supported by the National Key R\&D Program of China under Grant No. 2017YFB1003103; the National Natural Science Foundation of China under Grant Nos. 61300049, 61763003; and the Natural Science Research Foundation of Jilin Province of China under Grant Nos. 20180101053JC, 20190201193JC.
	
\section*{References}
	
\bibliography{mybibfile}

\end{document}